\documentclass[a4paper,twoside]{article}

\usepackage{epsfig}
\usepackage{subfig}
\usepackage{calc}
\usepackage{amssymb}
\usepackage{amstext}
\usepackage{amsmath}
\usepackage{amsthm}
\usepackage{multicol}
\usepackage{pslatex}
\usepackage{apalike}
\usepackage{paralist} 
\usepackage[table]{xcolor} 
\usepackage{threeparttable}
\usepackage{algorithm}
\usepackage[noend]{algpseudocode}
\usepackage{graphicx}
\usepackage{SCITEPRESS}     
\usepackage{url}

\begin{document}

\title{Revisiting Gray Pixel for Statistical Illumination Estimation}

\author{\authorname{Yanlin Qian\sup{1,3}, Said Pertuz\sup{1}, Jarno Nikkanen\sup{2}, Joni-Kristian K\"am\"ar\"ainen\sup{1} and Jiri Matas\sup{3}}
\affiliation{\sup{1}Laboratory of Signal Processing, Tampere University of Technology}
\affiliation{\sup{2}Intel Finland}
\affiliation{\sup{3}Center for Machine Perception, Czech Technical University in Prague}
\email{yanlin.qian@tut.fi}
}


\abstract{ We present a statistical color constancy method that relies on novel gray pixel detection and mean shift clustering. The method,  called Mean Shifted Grey Pixel -- MSGP, is based on the observation: true-gray pixels are aligned towards one single direction. Our solution is compact, easy to compute and  requires no training. 
Experiments on two real-world benchmarks show that the proposed approach outperforms state-of-the-art methods in the camera-agnostic scenario. In the setting where the camera is known, MSGP outperforms all statistical methods.
}

\onecolumn \maketitle \normalsize \vfill

\section{\uppercase{Introduction}}
\label{sec:introduction}

\begin{figure*}[t]
\begin{center}
\includegraphics[width=0.8\linewidth]{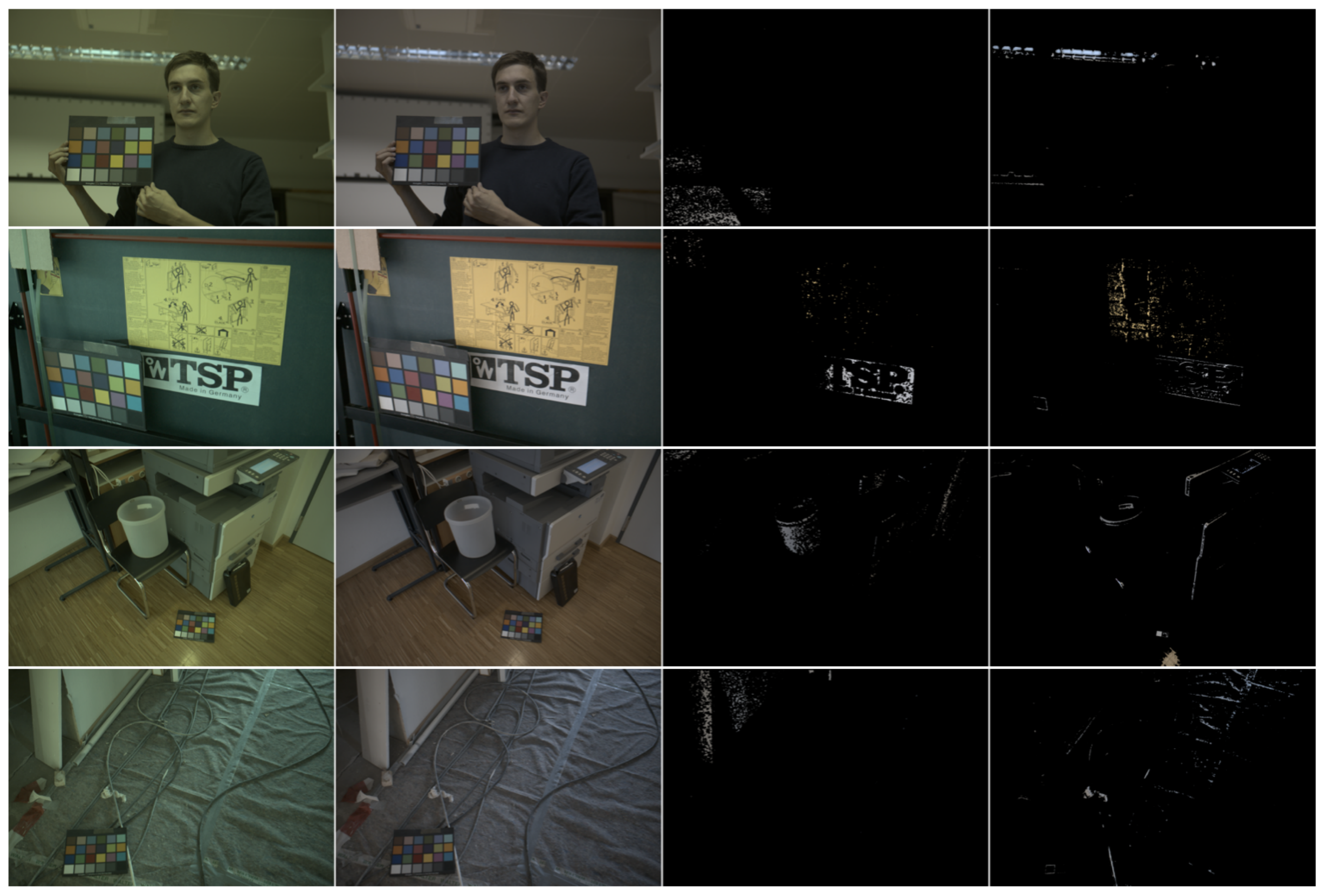}
\caption{Detection of gray pixels. From left to right: input image, color-corrected image using ground-truth, pixels chosen by the proposed method in Section \ref{subsec:clustering}, and pixels chosen by \cite{yang2015efficient}. Macbeth Color Checker are masked out due to both methods find gray pixels on gray regions.} 
\label{fig:comparison_twogp}
\end{center}
\end{figure*} 

\noindent The human eye automatically adapts to changes in imaging conditions and illumination of the scenes. Analogously, the ability of making color images look natural regardless of changing illumination is known as \textit{color constancy} and is an important feature of consumer digital cameras in order to yield visually canonical images.  
Color constancy is an important step in different computer vision applications, such as fine-grained classification, semantic segmentation, scene rendering and object tracking, among others \cite{foster2011color}.

For decades, the classical approaches for color constancy in digital cameras, \textit{statistical methods}, have relied on the assumption that some global or local statistical properties of the illumination are constant and can therefore be estimated directly from the image \cite{brainard1986analysis,barnard2002comparison,van2007edge,finlayson2004shades,gao2014eccv,yang2015efficient,Cheng14}. This approach has the advantage of being independent to the acquisition device since the properties of the scene illumination are estimated in a per-image basis. Recently, state-of-the-art methods including convolutional neural networks (CNN), namely \textit{learning-based methods} \cite{chakrabarti2012color,gijsenij2010generalized,gehler2008bayesian,gijsenij2011color,joze2014exemplar},  have consistently outperformed statistical methods when validated in several mainstream benchmarks. We argue that learning-based methods depend on the assumption that the statistical distribution of the illumination in both the training and testing images is similar. In other words, learning-based methods assume that imaging and illumination conditions of a given image can be inferred from previous training examples, thus becoming heavily dependent on the training data \cite{gao2017josa}.

In order to assess the limitation of color constancy methods to cope with differences between training and testing images, we focus on the \textit{Camera-agnostic color constancy} setting. For illustration, consider the case when a user retrieves an image from the unknown camera\footnote{We assume the image is with linear response and calibrated black offset, where color constancy method should be applied. Note that images over web are not usually this case.} and wants to color correct it. In this scenario, in which very little is known about the camera or capturing process of the image, color correction must be performed without strong assumptions on the source of the image or imaging device. In this less researched but still important setting, we experimentally show that, in camera-agnostic color constancy, learning-based methods perform poorly compared to statistical methods. As a result, there is a need for approaches that are insensitive to parameters such as the camera or imaging process used to capture the image. 

In this paper we propose a new statistical color constancy method. The proposed method, called \textit{mean-shifted gray pixel}, or MSGP, is a process that detects pixels that are assumed to be gray under neutral illumination. Why gray pixels? Gray or nearly gray pixels are wide spread in indoor and outdoor images~\cite{yang2015efficient}. In the process of manufacturing camera, each camera is calibrated to maintain: \textit{gray pixels will be rendered gray in linear image (not raw response) under standard neutral illumination}. Gray pixel examples are shown in the third column in Fig. \ref{fig:comparison_twogp}.

Considering that gray pixels are informative \textit{w.r.t.} casting illumination, it is possible to transform the scene illumination estimation task into gray pixel detection. This paper proposes an accurate method for the detection of gray pixels by combining a novel \textit{grayness measure} with Mean-shift clustering in color space. 

Experimental results in camera-agnostic color constancy show that the proposed algorithm outperforms both statistical and learning-based methods of the state-of-the-art. Even in the non camera-agnostic scenario, \textit{i.e.} using k-fold cross validation in the same datasets, the proposed method outperforms other statistical methods and shows a competitive performance when compared to learning-based methods.

\section{\uppercase{Previous Related Work}}

\noindent Assuming a photometric linear image $I$ captured using a digital camera, with pixels below black level and above saturation level corrected,  the simplified imaging formation under one global illumination source can be expressed as \cite{gijsenij2011computational}: 
\begin{equation}
I_{i}(x,y) = \int L(\lambda)S_{i}(\lambda)R(x,y,\lambda) d\lambda, i\in 
  \{\mbox{R,G,B}\},
\label{eq:formation}
\end{equation}
where $I_{i}(x,y)$ is the measured image color value at spatial location $(x,y)$, $L(\lambda)$ the wavelength distribution of the global light source, $S_{i}(\lambda)$ the spectral response of the color sensor, $R(x,y,\lambda)$ the surface
reflectance and $\lambda$ the wavelength. 

Under the narrow-band assumption (Von Kries coefficient law \cite{von1902}), Eq. \ref{eq:formation} can be further simplified as (same as \cite{barron2015convolutional}):

\begin{equation}
I =  W L,
\label{eq:simpleformation}
\end{equation}
which shows that the whole captured image $I$ is the element-wise Hadamard product of the white-balanced image $W$ and the illumination $L$. 

The \textit{goal} of all color constancy methods, both learning-based and statistical methods, is to estimate $L$, so as to recover $W$, given $I$.

\noindent\textbf{Learning-based Methods} \cite{chakrabarti2012color,gijsenij2010generalized,gehler2008bayesian,gijsenij2011color,joze2014exemplar,yanlin2016icpr,yanlin2017iccv} aim at building a model that relates the captured image $I$ and the sought illumination $L$ from extensive training data. Among the best-performing state-of-the-art approaches, the CCC method discriminatively learns convolutional filters in a 2D log-chroma space \cite{barron2015convolutional}. This framework was subsequently accelerated using the Fast Fourier Transform on a chroma torus \cite{barron2017cvpr}. Chakrabarti \textit{et al.} \cite{chakrabarti2015color} leverage the normalized luminance for illumination prediction by learning a conditional chromaticity distribution.
DS-Net \cite{shi2016eccv} and FC$^4$ Net \cite{hu2017cvpr} are two representative methods using deep learning. The former network chooses an estimate from multiple illumination guesses using a two-branch CNN architecture, while the later addresses local estimation ambiguities of patches using a segmentation-like framework. Learning-based methods achieve great success in predicting pre-recorded ``ground-truth'' illumination color to a fairly high accurate level, but heavily depending on the same cameras being used in both training and testing images (see Sections \ref{sec:unseen} and \ref{sec:agnostic}). The Corrected-Moment method \cite{finlayson2013corrected} can also be considered as a learning-based method as it needs to train a corrected matrix for each dataset. 

\noindent\textbf{Statistical Methods} 
estimate illumination by making some assumptions about the local or global regularity of the illumination and reflectance of the input image. The simplest such method is \textit{Gray World} \cite{buchsbaum1980spatial}, that assumes that the global average of  reflectance is achromatic. The generalization of this assumption by restricting it to local patches and higher-order gradients has led to some classical and recent statistics-based methods, such as White Patch \cite{brainard1986analysis}, General Gray World \cite{barnard2002comparison}, Gray Edge \cite{van2007edge}, Shades-of-Gray \cite{finlayson2004shades} and LSRS \cite{gao2014eccv}, among others \cite{Cheng14}. The closest works to ours are Xiong \textit{et al.} \cite{xiong2007automatic} and Gray Pixel \cite{yang2015efficient}. Xiong \textit{et al.} \cite{xiong2007automatic} finds gray surfaces based on a special LIS space, but this method is camera-dependent. The Gray Pixel method will be discussed in Section~\ref{sec:greypixelmethod}.

\noindent\textbf{Physics-based and other Methods} \cite{tominaga1996multichannel,finlayson2001convex,finlayson2001solving} estimate illumination from the understanding of the physical process of image formation (\textit{e.g.} the Dichromatic Model), thus being able to model highlights and inter-reflections. Most physics-based methods estimate illumination based on intersection of multiple dichromatic lines, making them work well on toy images and images with only a few surfaces but not very reliable on natural images \cite{finlayson2001solving}. The latest physics-based method is \cite{sung2018tip}, which relies on the longest dichromatic line segment assuming Phong reflection model holds and an ambient light exists.  Although our method is based on the Dichromatic Model, we classify our approach as \textit{statistical} since the core of the method is finding gray pixels based on some observed image statistics. We refer readers to \cite{gijsenij2011computational} for more details about physics-based methods. 

\noindent The \textbf{contribution} of this paper is three-fold:
\begin{compactitem}
\item We experimentally demonstrate that, in the camera-agnostic color constancy setting, state-of-the-art learning-based methods are outperformed by statistical methods
\item We point out the hidden elongated pixel prior over indoor and outdoor color constancy datasets.
\item We present the Mean-shift-based Gray Pixel method, robustly searching dominant illumination (mode) and achieving state-of-the-art performance among competing training-free alternatives. Code will be released upon publication.
\end{compactitem}

\section{\uppercase{Camera-agnostic Color Constancy}}
\label{sec:unseen}

\noindent For a given camera, noted as $C$, Eq. \ref{eq:simpleformation} can be rewritten as:

\begin{equation}
I_{C}=  W_C L_{C},
\label{eq:simpleformation_camera}
\end{equation}
which indicates that both, the captured image $I_C$, the canonical image $W_C$ and the illumination $L_C$ that we need to estimate, are dependent on the camera type $C$. $W_C$ indicates that in canonical light, the images captured by different cameras of the same scene differ. 

The color constancy problem in learning-based methods can be stated as ${\tilde{L}_C}=f(w,I_C)$, where $\tilde{L}_C$ is the estimated illumination, and $f(w,\cdot)$ is the mapping to be learned with parameters $w$. The mapping $f(w, \cdot)$ can be embodied by various machine learning models or an ensemble of them. If the learning process for a particular dataset is guided by the distance (\textit{e.g.} angular error) between ${{\tilde{L}_C}}$ and $L_C$, $w$ will undoubtedly be biased by the particular characteristics of the camera $C$. In other words, the parameters of $f(w,\cdot)$ will be learned to be ``well-performing'' on a specific dataset that encompasses one or a few pre-selected cameras. With the massive modeling capability of some machine learning models (\textit{e.g.} regression trees and deep learning), the camera sensibility function of a bag of cameras can be modeled up to a high degree. In the literature, the validation of color constancy methods is customarily performed using k-fold cross-validation on the same dataset. As a result, this validation process favors learning-based methods and fails to assess their performance for color correction in images from an unknown camera \cite{gao2017josa}.  

In this work, we define \textit{camera-agnostic color constancy} as the problem of estimating the illumination $L_C$ of a color-biased image $I_C$ that has been captured by a camera $C$ of unknown properties. For learning-based methods, this implies that the input image $I_C$ has been captured by a camera not previously ``seen'' in the training process. Therefore, a rigorous validation process of color constancy algorithms should consider both, camera-agnostic and known-camera scenarios. By leveraging publicly available datasets, this can be achieved by training in one dataset and testing in other without overlapping cameras (see Section \ref{sec:experiments}). In contrast to learning-based methods, statistical methods have the advantage of adjusting the model in a per-image basis thus having the potential to implicitly deal with the camera-agnostic problem. 

\section{\uppercase{Mean-Shifted Gray Pixel}}
\label{sec:greypixelmethod}
\noindent The proposed mean-shifted gray pixel algorithm, or MSGP, is built on the assumption that achromatic pixels in the corresponding canonical image can be used to estimate the global illumination. Specifically, achromatic pixels are visually gray in the color corrected image. 
Yang \textit{et al.} \cite{yang2015efficient} claimed the mentioned assumption, and experimentally demonstrated the presence of detectable gray pixels in most natural scenes under white light. In this work, we further extend the concept of the Gray Pixel method by means of an adaptive method for the detection gray pixels that combines a new grayness function and mean-shift clustering.

\subsection{Original Gray Pixel (GP) Revisited }
\label{subsec:basicgreypixel}

In this section, we revisited the original Gray Pixel method \cite{yang2015efficient}, which is derived from a limited diffuse reflection model. 
Applying a log transformation to both sides of \eqref{eq:simpleformation}, we have:

\begin{equation}
\log(I_i^{(x,y)}) = \log(W_i^{(x,y)})+\log(L_i)
\label{eq:simpleformation_log}
\end{equation}

In a small enough local neighborhood, the illumination $L$ can be assumed as uniform under global illumination constrains. As a result, the application of a linear channel-wise local contrast operator $C\{\cdot\}$ (Laplacian of Gaussian, which we will use for the remainder of the paper) on \eqref{eq:simpleformation_log} yields:
\begin{equation}
C\{\log(I_i^{(x,y)})\} = C\{\log(W_i^{(x,y)})\}
\label{eq:simpleformation_delta}
\end{equation}
Eq. \eqref{eq:simpleformation_delta} indicates a well-known observation: the casting illumination is irrelevant to the channel-wise local contrast of a small local neighborhood \cite{geusebroek2001color}. It also means that regions with no contrast are useless for obtaining illumination cues. Following \cite{yang2015efficient}, with balanced R, G and B responses, the following condition must be met by gray pixels:
\begin{equation}
 C\{\log(I_R^{(x,y)})\} = C\{\log(I_G^{(x,y)})\} = C\{\log(I_B^{(x,y)})\} \neq 0.
 \label{eq:threedelta}
 \end{equation}
In practice, \eqref{eq:threedelta} does not hold strictly. As a result, it is necessary to propose a ``grayness'' measure in order to detect nearly gray pixels. For the sake of simplicity, let us define the local contrast of a log-transformed image pixel located at $(x,y)$ as $\Delta_i(x,y)=C\{\log(I_i^{(x,y)})\}$ with $i\in\{R,G,B\}$. In \cite{yang2015efficient}, the grayness measure of a pixel, $G(x,y)$, is defined as:
\begin{equation}
G(x,y) = \left( \frac{1}{3}\sum_{i\in\lbrace R,G,B \rbrace} 
\frac{(\Delta_i(x,y)-\bar{\Delta}(x,y))^2}{\bar{\Delta}(x,y)} \right)^{1/2},
 \label{eq:greyness_oldgreypixel}
 \end{equation}
where $\bar{\Delta}(x,y)$ is the average of channels R, G and B.

It is claimed that the smaller $G(x,y)$ is, the more gray a pixel is under white light. Then some post-processing steps are applied to weaken dark pixels (luminance as dominator) and isolated pixels (local averaging), for which we refer readers to the original GP \cite{yang2015efficient}.

A major drawback of Eq. \ref{eq:greyness_oldgreypixel} is that the grayness estimate depends on the luminance of the pixels.
Specifically, the effect of $\bar{\Delta}$ results in gray pixels having different grayness values due to differences in luminance. Alternatively, we propose that grayness should only depend on \textit{chromaticity}. Therefore, in the next section, we will introduce a new grayness function  to replace Eq. \ref{eq:greyness_oldgreypixel}. 

\subsection{Grayness Function}
\label{subsec:angulargrayness}
We propose an ideal grayness function $G(\cdot) \in [0,1]$ where $0$ denotes pure gray of a pixel color.  Without specification, the grayness function works in RGB space as it is closest to the image formation process and main choice of in line of research \cite{yang2015efficient,barron2017cvpr}. Our grayness function should comply with the following properties:

\begin{tabular}{p{0.2\linewidth}p{0.68\linewidth}}
\textbf{Property 1}&$G(\cdot)$ is invariant to the luminance (sum of RGB values).\\
\textbf{Property 2}&$G(\cdot)$ outputs monotonically decreasing value for increasing visual grayness, \textit{e.g.} from red to white.\\
\textbf{Property 3}&Pure gray pixels (on the black-to-white line) should be have value $0$.
\end{tabular}

In addition the three above-mentioned properties, it is also desirable that the output space of the grayness function be normalized (so that no subsequent normalization is required), as well as having a physical meaning so that it can be used for other computer vision tasks. Alternatively to the grayness measure proposed in \cite{yang2015efficient}, we propose a new grayness measure based on the angular error function that complies with all these properties:
\begin{equation}
G(x,y)=\cos^{-1} \left( \frac{\langle  {\Delta}(x,y),\textbf{g}\rangle}
{\Vert{ {\Delta}(x,y)}\Vert\Vert{\textbf{g}}\Vert_2}  \right),
\label{eq:grayness_newgreypixel}
\end{equation}
where ${\Delta}(x,y) = [\Delta_r, \Delta_g, \Delta_b]^\intercal$ is the RGB vector in location $(x,y)$, $\textbf{g}$ is the gray light reference vector $[g_r, g_g, g_b]^\intercal$, and $\Vert \cdot \Vert_n$ refers to the $\ell n$ norm.

Our motivation behind Eq. \ref{eq:grayness_newgreypixel} is that, even in the color-biased scenario, it is possible to assume that all gray colors captured by the same camera will have balanced R, G, B components, regardless of their luminance level. As a result, it is possible to assess their grayness level by measuring the angular error with respect to a reference gray value. Notice that, in general, the gray reference vector $\textbf{g}$ can have spatially-varying values in order to adjust for changes in the illumination of the scene. In this work, however, we assume that the global illumination source remains constant in the scene and adopt the canonical gray value as reference: $\textbf{g} = [1, 1, 1]^\intercal$. In this case, Eq. \ref{eq:grayness_newgreypixel} can be further simplified as:
\begin{equation}
G(x,y)=\cos^{-1} \left(\frac{1}{\sqrt{3}} \frac{\Vert{ {\Delta}(x,y)}\Vert_1}
{\Vert{ {\Delta}(x,y)}\Vert_2}  \right),
\label{eq:grayness_newgreypixel2}
\end{equation}

Eq. \ref{eq:grayness_newgreypixel2} measures how gray a pixel is, using the angular distance from the local contrast vector to the gray light $\textbf{g}$, thus meeting Properties 1 and 2. When the point $(x,y)$ is completely gray, $G(x,y)$ is $0$ and increases monotonically with decreasing level of grayness, thus meeting Property 3. In addition, the output ranges from $0^\circ$ to $\cos^{-1}(\frac{1}{\sqrt{3}})$ for each image, thus being normalized. 

\begin{figure}[hbt]
\begin{center}
{\includegraphics[width=.2\linewidth]{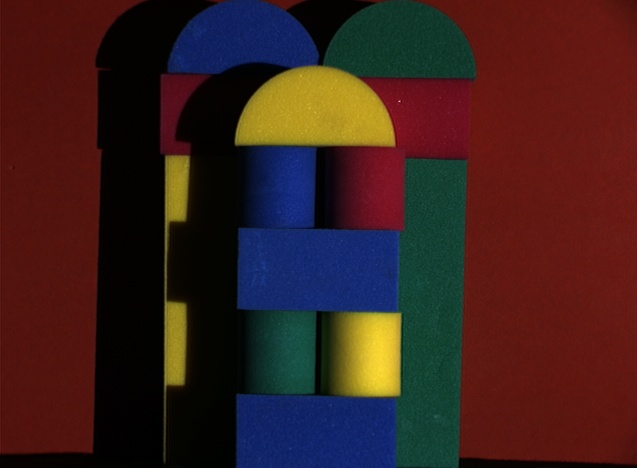}\label{fig:no_cc_1}}\hspace{.01cm}
{\includegraphics[width=.2\linewidth]{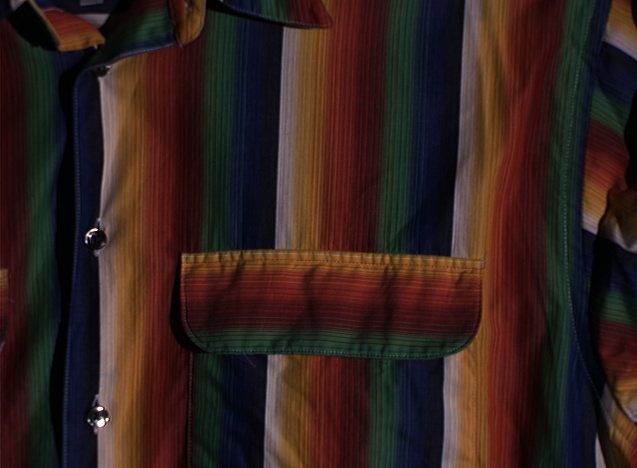}\label{fig:no_cc_2}}\hspace{0.01cm}
{\includegraphics[width=.2\linewidth]{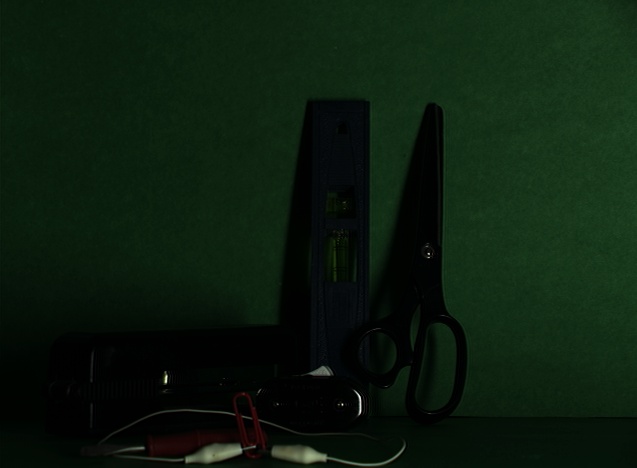}\label{fig:no_cc_3}}\hspace{.01cm}
{\includegraphics[width=.2\linewidth]{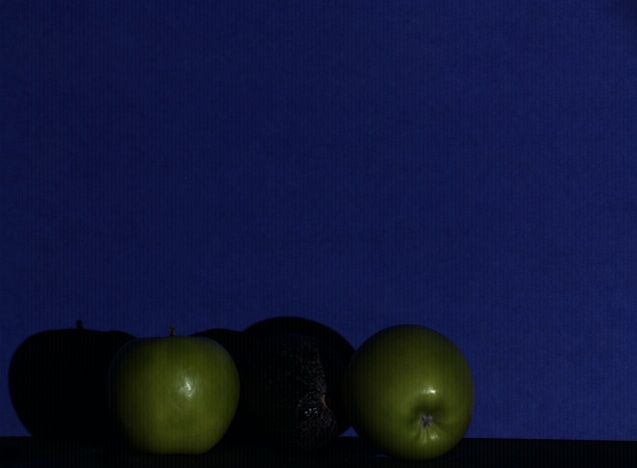}\label{fig:no_cc_4}}\hspace{0.01cm}

\caption{Examples of SFU Indoor dataset}
\label{fig:indoorcases}
\end{center}\vspace{-0.1cm}
\end{figure}

\vspace{\medskipamount}\noindent\textbf{Empirical Evidence -- }The next question is whether this new grayness function brings different ordering of pixels according to their grayness levels. To answer this, we replace Eq. \ref{eq:greyness_oldgreypixel} with Eq. \ref{eq:grayness_newgreypixel} in the original GP algorithm and estimate illumination in two mainstream color constancy benchmarks where GP is evaluated. Table \ref{tab:grayness} shows the performance improvement by a large margin ($0.6^\circ$ reduction in median error for SFU Color Checker) when we use the proposed grayness measure Eq. \ref{eq:grayness_newgreypixel}. Results on the SFU Indoor dataset do not differ much, arguably because the dataset is collected in a laboratory environment with a restricted set-up (many image feel artificial and examples are shown in Fig \ref{fig:indoorcases}). The proposed method is based on the assumption of natural image statistics and works for more general cases. 
For the results shown in Table \ref{tab:grayness}, the top $0.1\%$ pixels with $G$ values are chosen as gray pixels, as recommended by \cite{yang2015efficient}. The local contrast operator $C\{\cdot\}$ is the Laplacian of Gaussian.

\begin{table}[ht]
\small
\caption{Angular error of the Gray Pixel ~\cite{yang2015efficient} algorithm with different grayness functions: original grayness function (GP) and proposed grayness function in Eq. \ref{eq:grayness_newgreypixel2} (GP$^\ast$)} 
\label{tab:grayness}
\begin{center}
\scriptsize
  \begin{tabular}{l rrr rrr}
\hline
  &  \multicolumn{3}{c}{{SFU Color Checker}}  & \multicolumn{3}{c}{{SFU Indoor}}\\
  & Mean & Med & Trimean & Mean & Med & Trimean\\
\hline
GP & 4.6 &  3.1 & --  & 5.3 & 2.3 & -- \\
GP$^\ast$ & 4.1 &  2.5 & 2.8  & 5.3 & 2.2 & 2.7  \\ 
\hline
  \end{tabular}
\end{center}\vspace{-0.1cm}
\end{table}

\vspace{-1ex}

Here we mathematically analyze the connection between the grayness function in Eq. \ref{eq:greyness_oldgreypixel} and the proposed grayness measure in Eq. \ref{eq:grayness_newgreypixel2}. To avoid readers' confusion, we term the original grayness measure in Eq. \ref{eq:greyness_oldgreypixel} as $G_{\sigma}(x,y)$ and the proposed grayness function of Eq. \ref{eq:grayness_newgreypixel} as $G_\theta(x,y)$. In the sequel, we demonstrate that $G_\sigma$ and $G_\theta$ are related by:

\begin{equation}
G_{\sigma}(x,y)=\gamma(x,y) G_{\theta}(x,y),
\label{eq:g_connection0}
\end{equation}
where $\gamma(x,y)$ is a luminance-dependent term.

In order to demonstrate the relationship in Eq. \ref{eq:g_connection0}, we approximate $G_\theta$ as follows\footnote{For the sake of simplicity we will drop the pixel coordinates $(x,y)$ in the remaining of this section}:
\begin{equation}
G_{\theta}\approx\sqrt{1-\frac{1}{\sqrt{3}} \frac{\Vert{\Delta}\Vert_1}
{\Vert{\Delta}\Vert_2} }
\label{eq:grayness_cosinedistance}
\end{equation}

It can be readily shown that Eq. \ref{eq:grayness_cosinedistance} is an approximation of the same order of Eq. \ref{eq:grayness_newgreypixel2} in the interval $[0, 1]$. With this approximation, $G_\theta$ and $G_\sigma$ can be rewritten as:

\begin{align}
3\beta G_{\sigma}^2 &= {\alpha^2}-3\beta ^2\label{eq:g_beta}\\
G_{\theta}^2 &= 1-\frac{\sqrt{3}\beta}{\alpha}\label{eq:g_alpha}
\end{align}
where $\alpha = \Vert{\Delta}\Vert_2$ and $\beta =\frac{1}{3} \Vert{\Delta}\Vert_1$.

Putting a multiplier $\alpha(\alpha+\sqrt{3}\beta)$ to both sides of Eq. \ref{eq:g_alpha} yields: 
 
\begin{equation}
\alpha(\alpha+\sqrt{3}\beta) G_{\theta}={\alpha^2}-3\beta ^2
\label{eq:g_cos_multiplier}
\end{equation}

Finally, combing Eq. \ref{eq:g_beta} and \ref{eq:g_cos_multiplier} we obtain the sought relationship:
 
\begin{equation}
G_{\sigma}=\gamma G_{\theta},
\label{eq:g_connection}
\end{equation}
where $\gamma^2$ equals to $\alpha(\alpha+\sqrt{3}\beta)/3\upsilon$.
 
From Eq. \ref{eq:g_connection} it is clear that the original grayness function $G_{\sigma}(x,y)$ contains not only the real grayness -- cosine distance $G_{\theta}(x,y)$ from the gray light -- but also introduces a non-linear luminance-dependent term $\gamma(x,y)$, which adds noise to the grayness estimate. As a result, two points with same values of $G_{\theta}(x,y)$ but different luminance values will yield different values of $G_{\sigma}(x,y)$. In contrast, the proposed grayness function $G_{\theta}(x,y)$ is more robust to changes in luminance.
 
After some post-processing steps (\textit{e.g.} local averaging and normalization by image intensity), a small percentage of pixels ($N\%$) with the highest grayness values (lowest $G$) are chosen and averaged to be the illumination estimate. However, as it will be shown in the next section, the chosen gray pixels may still contain a number of colorful pixels. As a result, we will apply Mean Shift clustering in 3D RGB space in order to remove spurious color pixels. In the experiments in the remaining of this paper, we will use the new grayness function unless indicated otherwise. 

\subsection{Mean Shift Purification}
\label{subsec:clustering}

Let $S$ be the set of preselected $N$\% pixels according to their grayness levels. Ideally, $S$ should only contain pure-gray pixels. However, in fact $S$ may contain a number of colorful pixels that need to be removed before estimating the global illumination of the scene. 

\begin{figure}
\begin{center}

\subfloat[]{\includegraphics[width=.32\linewidth]{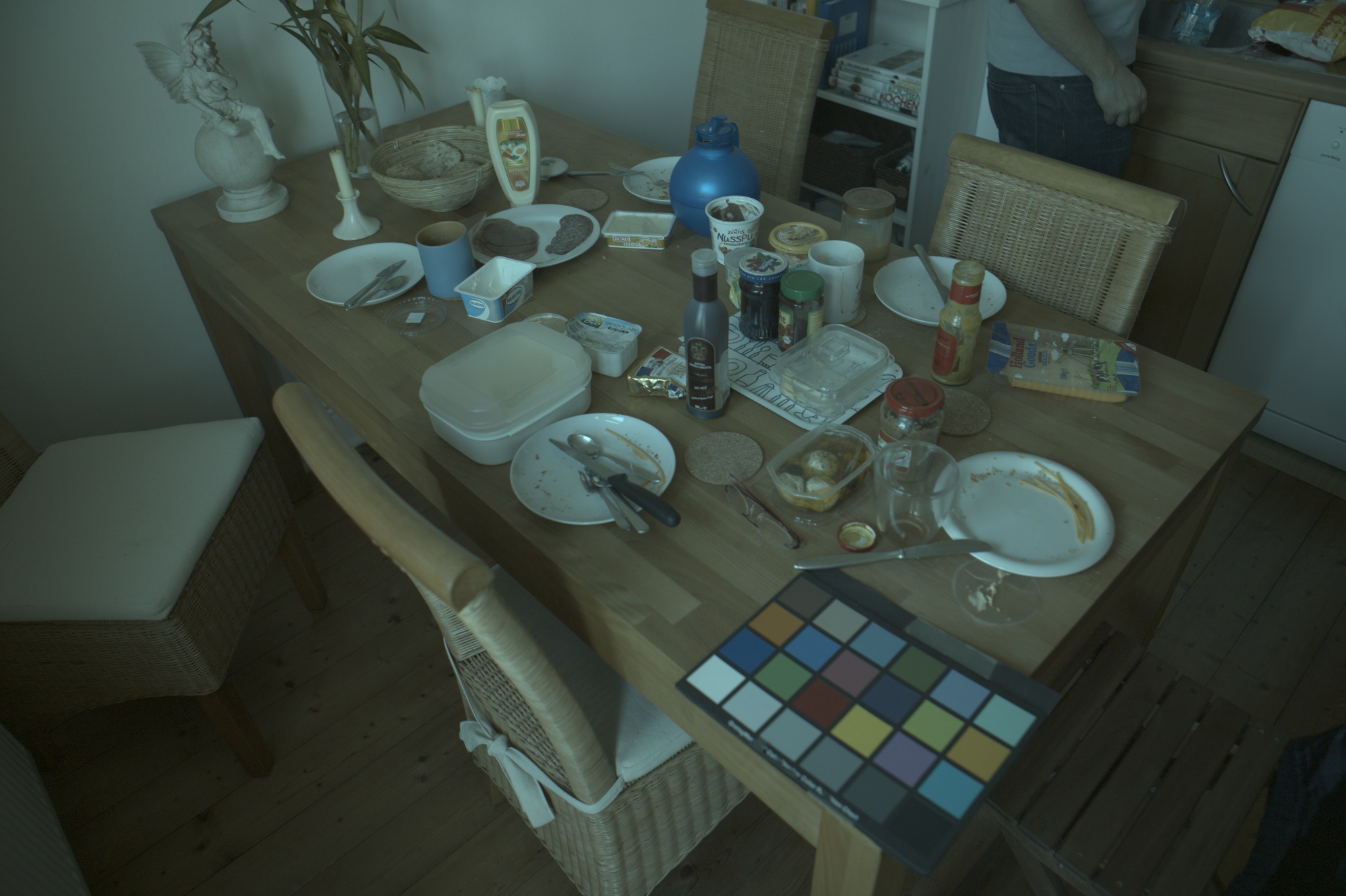}\label{fig:sfig1_1}}\hfill  
\subfloat[]{\includegraphics[width=.32\linewidth]{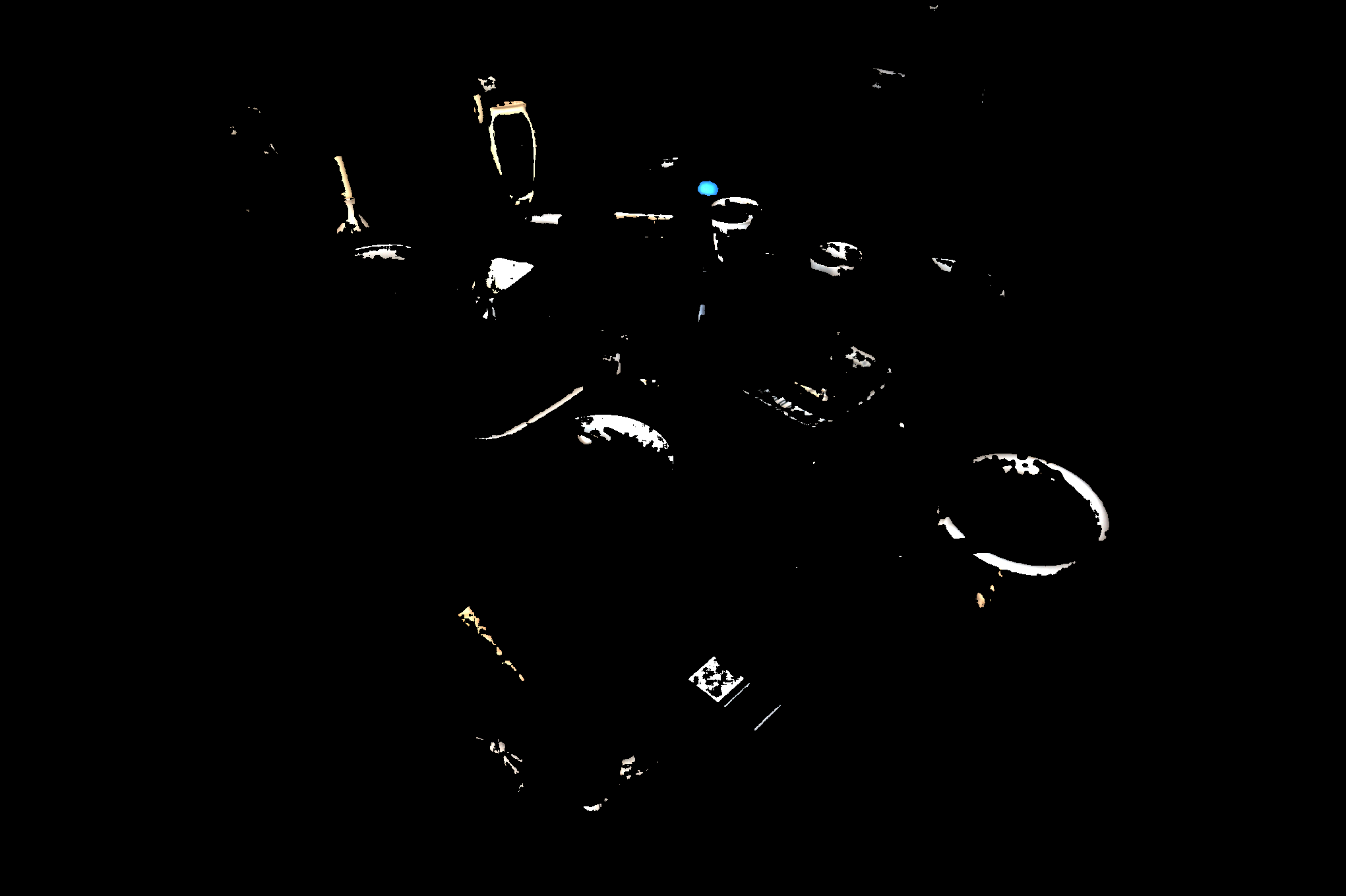}\label{fig:sfig1_2}}\hfill 
\subfloat[]{\includegraphics[width=.32\linewidth]{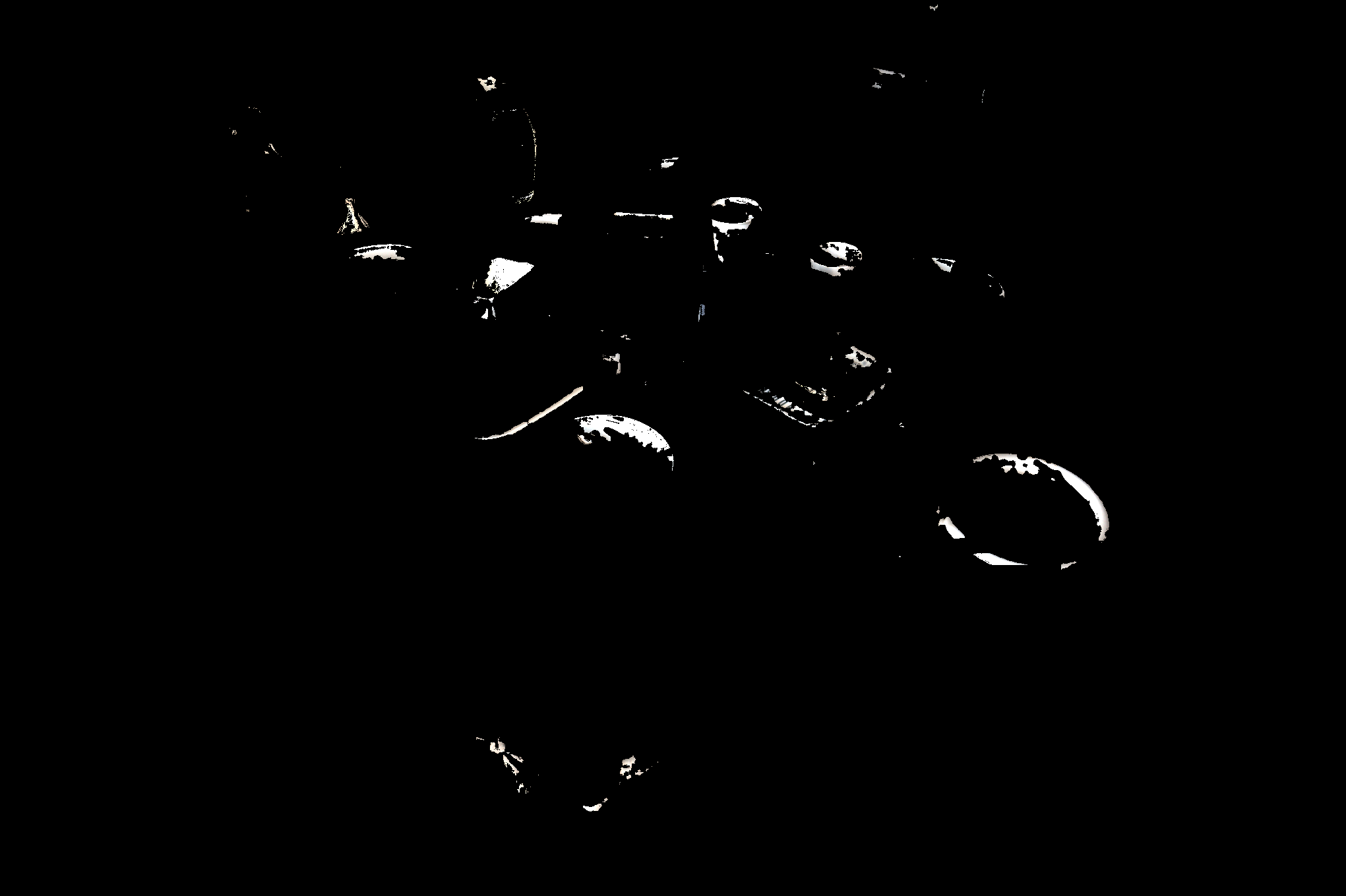}\label{fig:sfig1_3}}\\

\subfloat[]{\includegraphics[width=.33\linewidth]{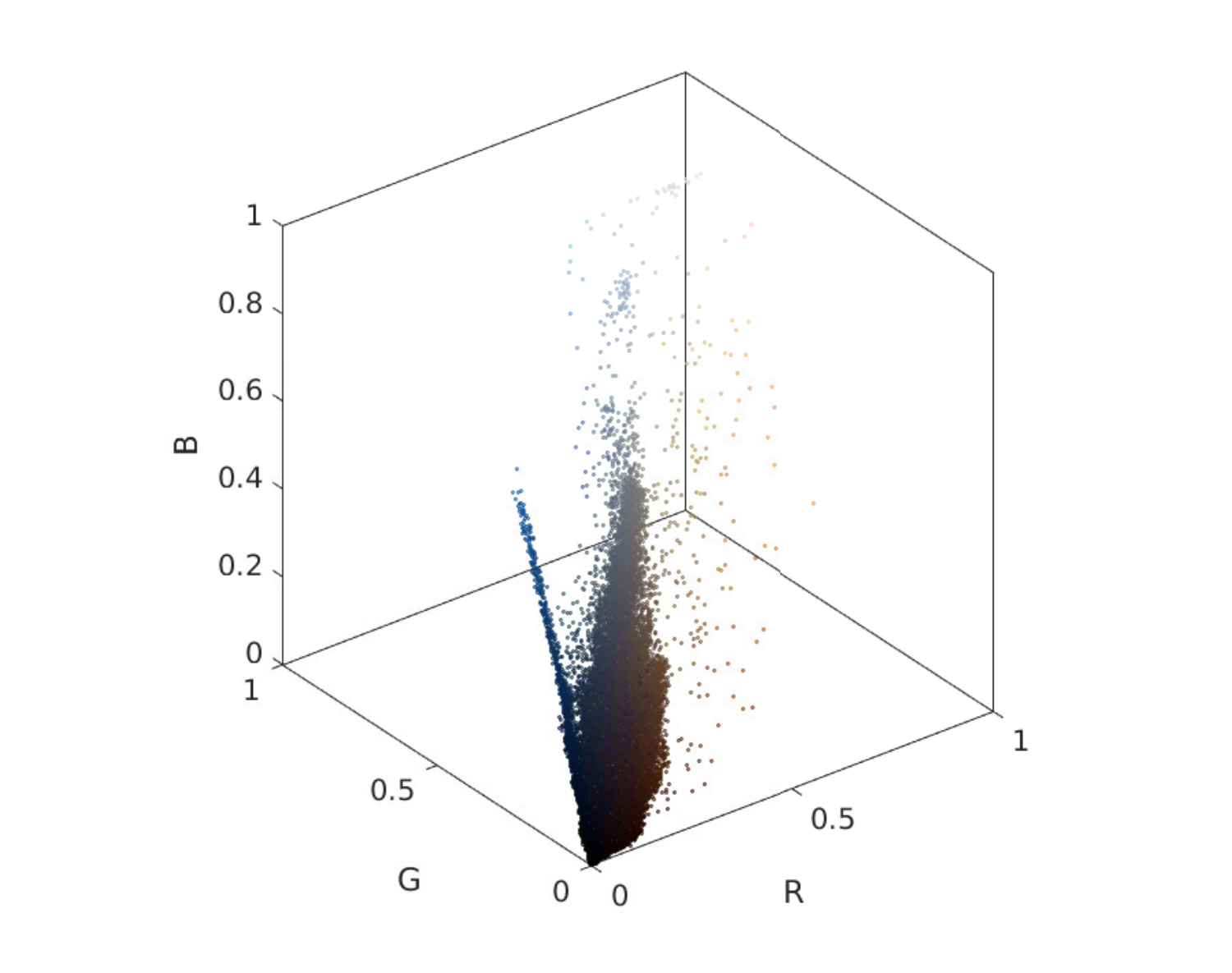}\label{fig:sfig1_4}}\hfill
\subfloat[]{\includegraphics[width=.33\linewidth]{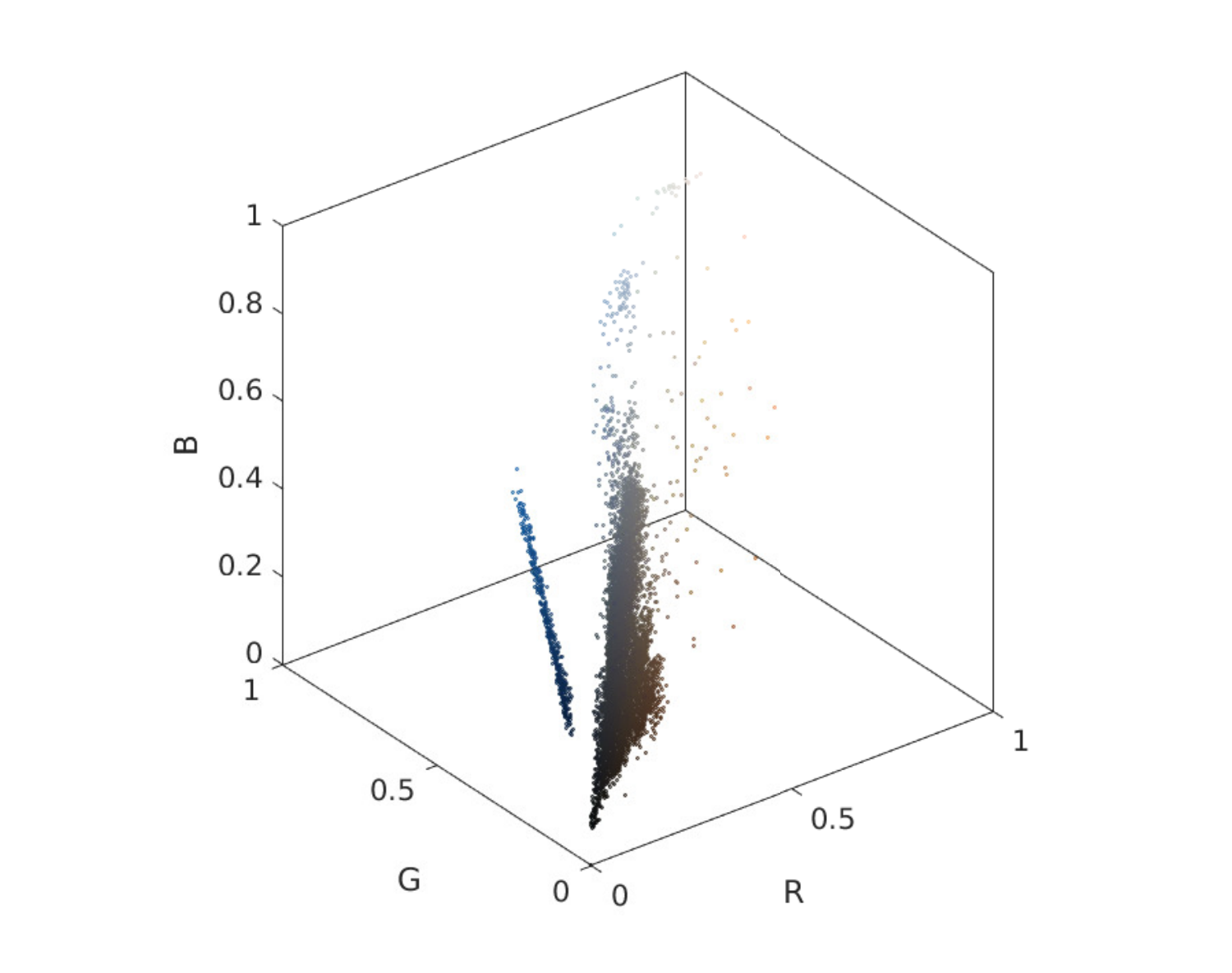}\label{fig:sfig1_5}}\hfill
\subfloat[]{\includegraphics[width=.33\linewidth]{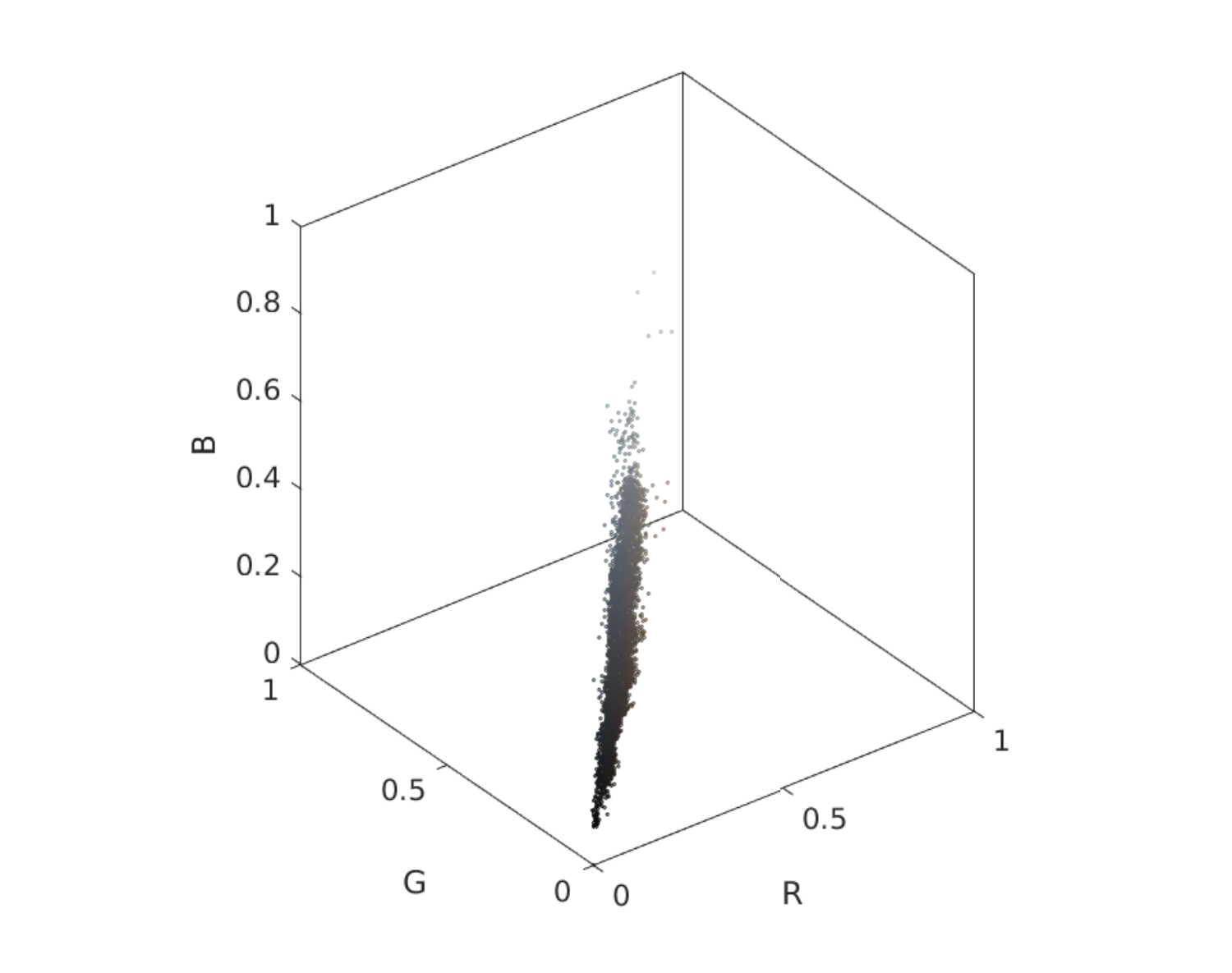}\label{fig:sfig1_6}}

\subfloat[]{\includegraphics[width=.32\linewidth]{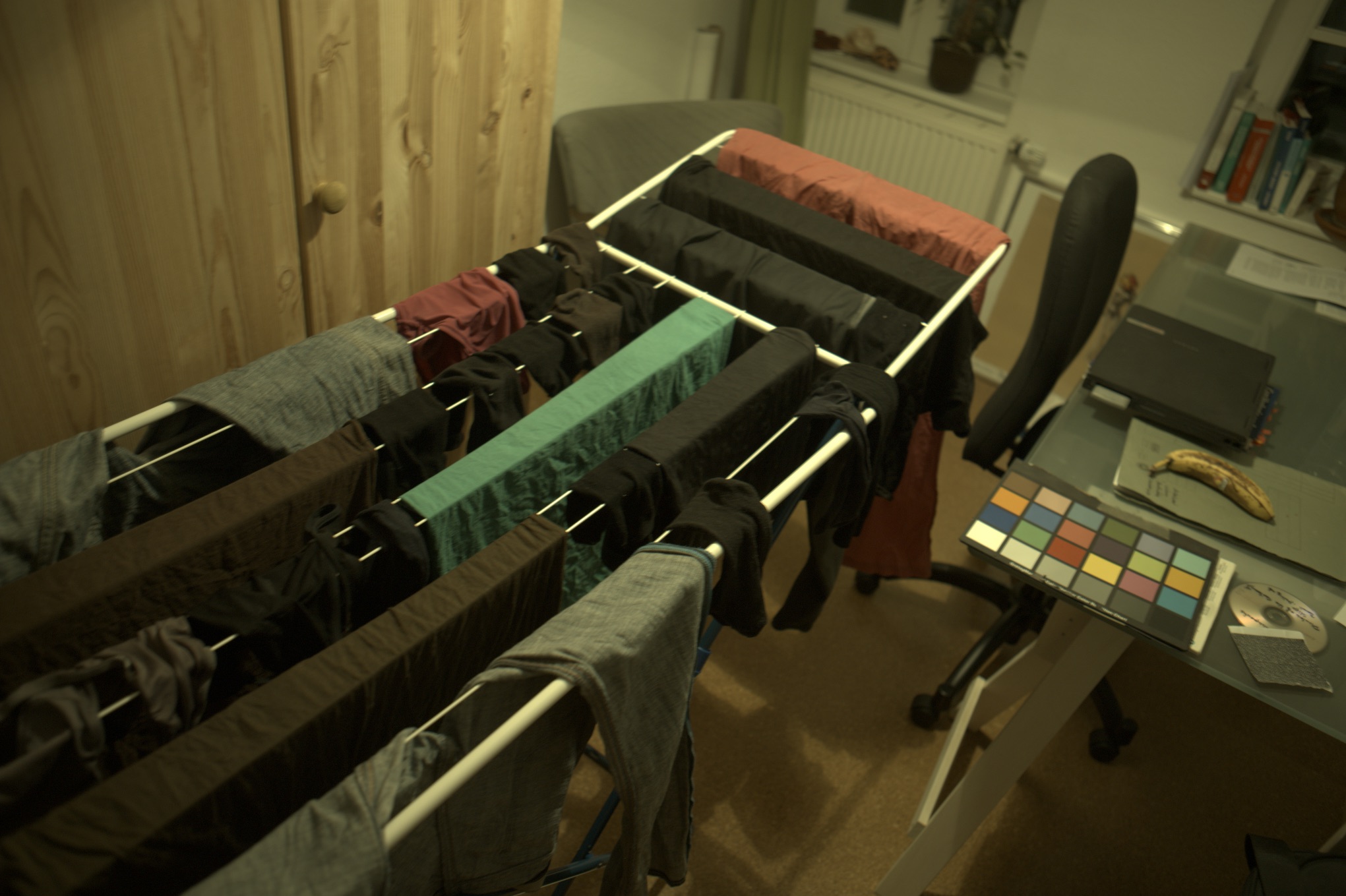}\label{fig:sfig2_1}}\hfill  
\subfloat[]{\includegraphics[width=.32\linewidth]{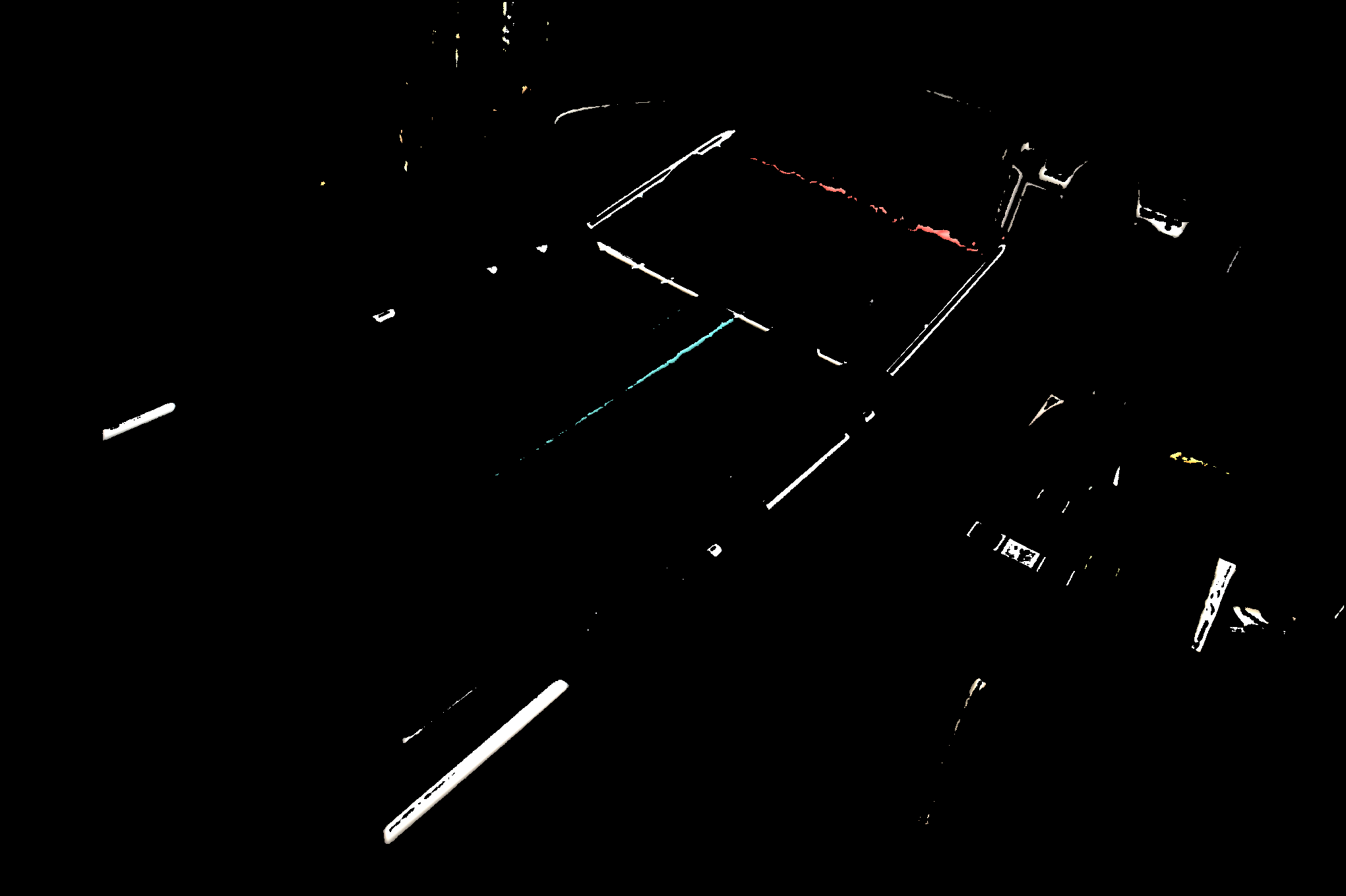}\label{fig:sfig2_2}}\hfill 
\subfloat[]{\includegraphics[width=.32\linewidth]{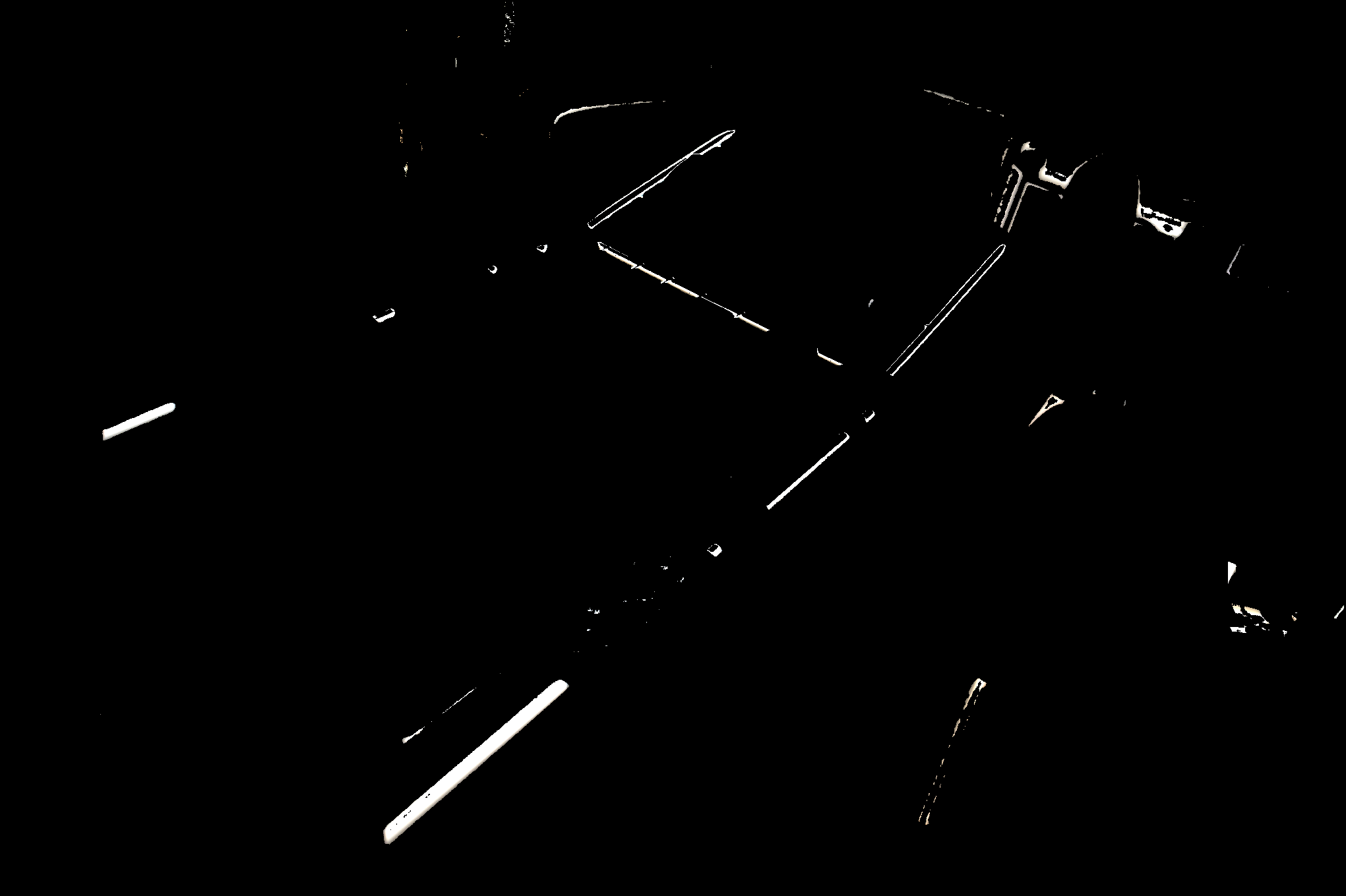}\label{fig:sfig2_3}}\\

\subfloat[]{\includegraphics[width=.33\linewidth]{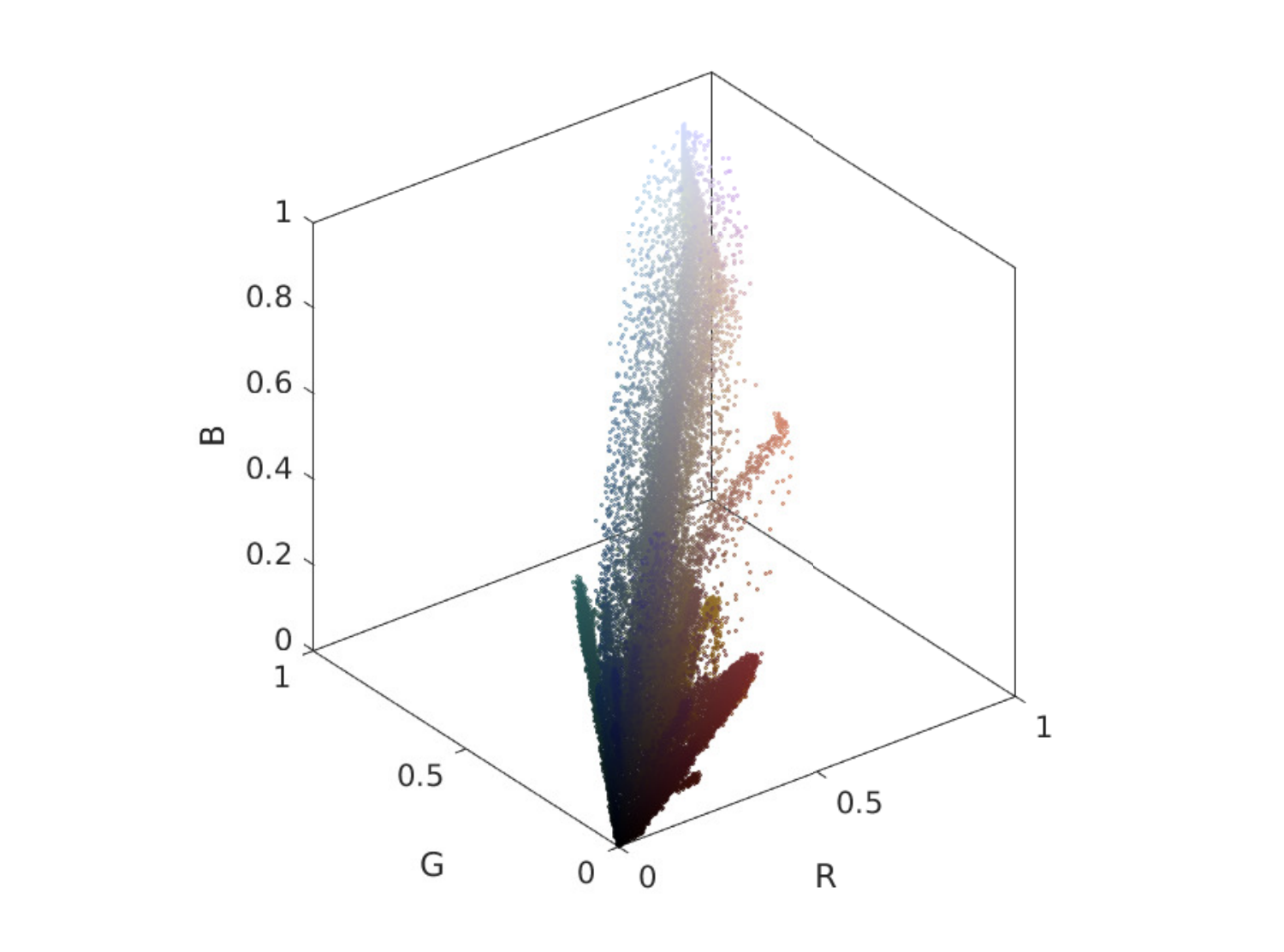}\label{fig:sfig2_4}}\hfill
\subfloat[]{\includegraphics[width=.33\linewidth]{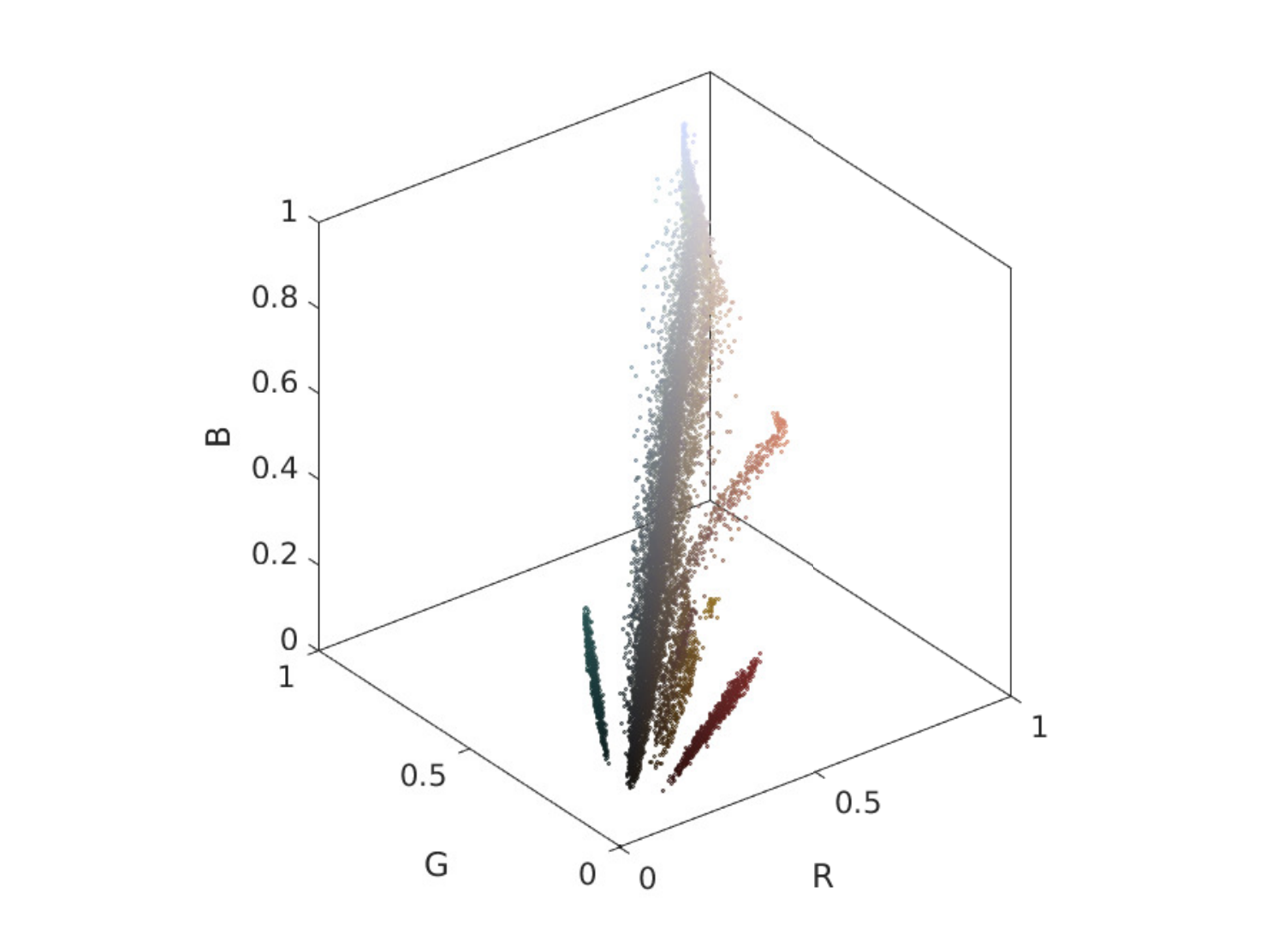}\label{fig:sfig2_5}}\hfill
\subfloat[]{\includegraphics[width=.33\linewidth]{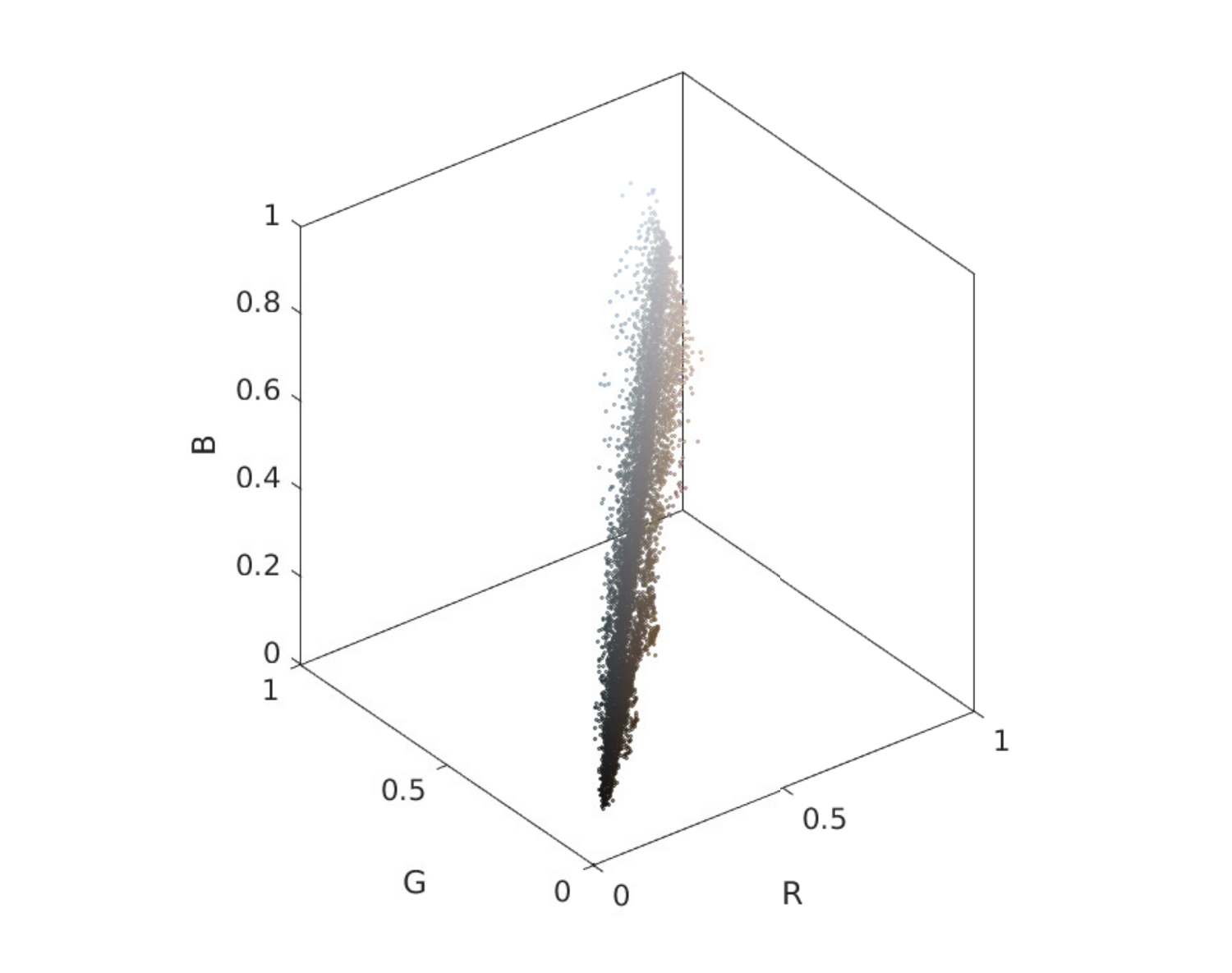}\label{fig:sfig2_6}}

\caption{Detection of gray pixels. After correction using ground-truth illumination, ideal gray pixel should looks purely gray. (a,g) Input image, (b,h) Initial gray pixels detected. (c,i) Purified gray pixels after the Mean Shift step. (d-f, j-l) color histograms of (a-c). Comparing (e) with (f), (j) with (l), it is clear that Mean Shift helps to discard color pixels in (e) that are not aligned with the main illumination vector. For visualization purposes, the luminance of (b,c) is multiplied by a constant $4$.}
\label{fig:cluster_examples}
\end{center}\vspace{-0.6cm}
\end{figure}

In order to remove color pixels from $S$, we note that, for a color-biased image $I$, all the pure-gray pixels should be contained in the illumination direction $[L_r,L_g,L_b]$. This is equivalent to having all the pixels aligned towards the gray-light vector $\textbf{g} = [1, 1, 1]^\intercal$ in the canonical image. For illustration purposes, Fig.~\ref{fig:sfig2_4} shows all the pixels of the canonical image of Fig.~\ref{fig:sfig2_1} in RGB space and Fig.~\ref{fig:sfig2_5} shows the corresponding set $S$ of pre-selected gray-pixels. From Fig.~\ref{fig:sfig2_5}, it is clear that $S$ contains both color and gray pixels. As predicted by our assumption, most true-gray pixels are aligned towards one single direction. In particular, the main direction of the densest pixel cloud indicates the illumination of the scene.

In this paper, we use \textit{mean shift} (MS) clustering \cite{ms1975,comaniciu2002mean} with a hybrid distance to seek for the dark-to-bright elongated cluster which contains the most pixels in $S$. MS is a non-parametric space analysis algorithm, treating the feature space as a probability density function and seeking for the modes. In this work, the density of each pixel $p\in S$ in RGB space is calculated as a function of the bandwidth $h$: 
\begin{equation}
\hat{f}(p)=\frac{1}{n}\sum_{i=1}^{n} K(p,p_i; h),
\label{eq:densityfunction}
\end{equation}
where $n$ the number of pixels in $S$, and the kernel density function $K(\cdot)$ is defined as: 
\begin{equation}
K(p, p_i; h) =
\begin{cases}
    1,& \text{if }  D(p,p_i) \leq h\\
    0,& \text{otherwise}
\end{cases}
\label{eq:kernel}
\end{equation}
with $\textbf{I}(p)=[I_r, I_g, I_b]$ being the vector with RGB values of pixel $p$, $D(p,p_i)$ is the defined hybrid distance computed as the product of the euclidean and angular distances $ \Vert \textbf{I}(p)-\textbf{I}(p_i)\Vert_2 \cdot
\angle\{\textbf{I}(p), \textbf{I}(p_i)\}$, and $\angle\{\cdot\}$ is the angle between two vectors.

Finally, the centroid corresponding to the mode with highest density is used for the computation of the illumination estimate:
\begin{equation}
\hat{L}=\mathop{\arg\max}_{p \in S} \hat{f}(p).
\label{eq:biggestdensity}
\end{equation}
The effect of mean shift clustering on the detection of gray pixels is illustrated in Fig.~\ref{fig:cluster_examples}. Comparing Figs.~\ref{fig:sfig2_2} and \ref{fig:sfig2_3}, it is clear how the mean-shift clustering, simply and effectively, allows for the detection and removal of color pixels in the initial set $S$. It is worthy to mention that, in some cases, there is almost no colored pixels in $S$. Fortunately, the performance will not suffer from clustering, as MS gracefully generates only one cluster which gives us a reliable estimate. As a result, there is no need to condition when to apply clustering.

The mean-shifted gray pixel algorithm (MSGP) is summarized in Algorithm \ref{alg:ss}. The proposed method depends only in two parameters: the percentage of pixels chosen from their grayness values, $N\%$, and the clustering bandwidth $h$ of Eq. \ref{eq:densityfunction}. The selection of these parameters and their effect on the performance of the proposed MSGP algorithm are presented in section \ref{sec:params}.

\begin{algorithm}[ht]
\caption{Mean-Shifted Gray Pixel}
\label{alg:ss}
\begin{algorithmic}[0]
\\
\textbf{Inputs:}
\State \indent $I$ \Comment{Color-biased image}\\
\textbf{Parameters:}
\State \indent $N$\Comment{Percentage of pixels}\\
\State \indent $h$\Comment{Bandwidth for MS clustering}\\
\textbf{Output:}
\State \indent $\hat{L}$  \Comment{Estimated illumination.}
\\
\\
\hrule \\
\textbf{Steps:}
\State 1. Compute local contrast $\Delta(x,y)$ 
\State 2. Compute grayness measure $G_\theta$(x,y). \Comment{Eq. \ref{eq:grayness_newgreypixel}}
\State 3. Generate $S$ with the top-$N\%$ gray pixels.
\State 4. MS clustering on $S$ with bandwidth $h$.\Comment{Eq. \ref{eq:densityfunction}}
\State 5. Select $\hat{L}$ as the strongest mode of $\hat{f}$. \Comment{Eq. \ref{eq:biggestdensity}}
\end{algorithmic}
\end{algorithm}
\vspace{-0.6cm}

\section{\uppercase{Experiments}}
\label{sec:experiments}
\noindent Experiments were conducted in two widely known, publicly available datasets collected for the purpose of evaluation of color constancy methods:
\begin{compactitem}
\item Gehler-Shi Dataset \cite{shi2010re}: $568$ high dynamic linear images, $2$ cameras \footnote{cameras: Canon 1D, Canon 5D}.
\item NUS 8-Camera Dataset \cite{Cheng14}: $1736$ high dynamic linear images, $8$ cameras \footnote{cameras: Canon 1DS Mark3, Canon 600D, Fujifilm X-M1, Nikon D5200, Olympus E-PL6, Panasonic Lumix DMC-GX1, Samsung NX2000, Sony SLT-A57}.
\end{compactitem}

The parameters of the proposed MSGP algorithm were selected as follows: the local contrast operator used in Eq. \ref{eq:grayness_newgreypixel2} was the Laplacian of Gaussian with a range of $5$ pixels. The bandwidth for MS clustering was set to $h=0.001$. The percentage of pixels chosen for the generation of $S$ was set to $N=0.1\%$. These parameters were selected based on preliminary experiments (see Section~\ref{sec:params}) and remained fixed for all the experiments. 

In order to allow for a rigorous comparison with state-of-the-art methods, we have considered two scenarios. The \textit{camera-agnostic} setting and the \textit{camera-known} setting. In the agnostic-camera setting, learning-based algorithms are trained in one dataset (\text{e.g.}, Gehler-Shi) and tested on the other. This allows for testing the performance of the algorithm in cameras not previously ``seen'' in the training process. The camera-known setting corresponds to the typical 3-fold cross validation used in the literature, in which learning-based methods are trained and validated in the same dataset. Visual comparison is given in Fig. \ref{fig:comparison_twogp}, where the proposed method detects gray pixels more accurately.
Numerical statistics are summarized in Table~\ref{tab:maintable} and discussed in Sections \ref{sec:agnostic} and \ref{sec:known}.

\begin{table*}
\caption{Quantitative Evaluation of CC methods. All values correspond to angular error in degrees. We report the results of the related work in the following order: 1) the cited paper, 2) Table [1] and Table [2] from Barron \textit{et al.}~\cite{barron2017fourier,barron2015convolutional} considered to be up-to-date and comprehensive, 3) the color constancy benchmarking website~\cite{colorconstancy.com}. We left dash on unreported results. In (a) results of learning-based methods worse than ours are marked in gray. The training time and testing time are reported in seconds, averagely per image, if reported in the original paper.   
}
\label{tab:maintable}

\begin{threeparttable}
\centering
\subfloat[single-dataset setting]{
\label{tab:known}
\resizebox{1.0\linewidth}{!}{
\begin{tabular}{l  rr rrr || rr rrr }

\hline
    & \multicolumn{5}{c}{Gehler-Shi} & 
    \multicolumn{5}{c}{{NUS 8-camera}}\\    
    
  & { Mean} & { Median} & Trimean & Best 25\% & Worst 25\% &{ Mean} & { Median} & Trimean & Best 25\% & Worst 25\% \\
\hline
\multicolumn{5}{c}{{\em Learning-based Methods (camera-known setting)}} \\
Edge-based Gamut ~\cite{gijsenij2010generalized}
 & \cellcolor{gray!25}6.52 & \cellcolor{gray!25}5.04 & \cellcolor{gray!25}5.43 & \cellcolor{gray!25}1.90 & \cellcolor{gray!25}13.58&  \cellcolor{gray!25}4.40  & \cellcolor{gray!25}3.30  & \cellcolor{gray!25}3.45 & \cellcolor{gray!25}0.99 & \cellcolor{gray!25}9.83\\
 Pixel-based Gamut~\cite{gijsenij2010generalized}
 & \cellcolor{gray!25}4.20 & \cellcolor{gray!25}2.33 & \cellcolor{gray!25}2.91 & \cellcolor{gray!25}0.50 & \cellcolor{gray!25}10.72 &  \cellcolor{gray!25}5.27  & \cellcolor{gray!25}4.26  & \cellcolor{gray!25}4.45 & \cellcolor{gray!25}1.28 & \cellcolor{gray!25}11.16 \\
Bayesian \cite{gehler2008bayesian}
& \cellcolor{gray!25}4.82 & \cellcolor{gray!25}3.46  &  \cellcolor{gray!25}3.88 & \cellcolor{gray!25}1.26 & \cellcolor{gray!25}10.49 & \cellcolor{gray!25}3.50  & \cellcolor{gray!25}2.36 & \cellcolor{gray!25}2.57 & \cellcolor{gray!25}0.78 & \cellcolor{gray!25}8.02\\
Natural Image Statistics~\cite{gijsenij2011color}
  & \cellcolor{gray!25}4.19 & \cellcolor{gray!25}3.13 & \cellcolor{gray!25}3.45 & \cellcolor{gray!25}1.00 & \cellcolor{gray!25}9.22& \cellcolor{gray!25}3.45 &  \cellcolor{gray!25}2.88  & \cellcolor{gray!25}2.95 &\cellcolor{gray!25}0.83 & \cellcolor{gray!25}7.18 \\
  
 Spatio-spectral (GenPrior) \cite{chakrabarti2012color}
 & \cellcolor{gray!25}3.59  & \cellcolor{gray!25}2.96& \cellcolor{gray!25}3.10 & \cellcolor{gray!25}0.95 & 7.61 &\cellcolor{gray!25}3.06  & \cellcolor{gray!25}2.58  & \cellcolor{gray!25}2.74 & \cellcolor{gray!25}0.87 & 6.17 \\

Corrected-Moment\tnote{1} (19 Edge)  \cite{finlayson2013corrected}
&  \cellcolor{gray!25}3.12 & \cellcolor{gray!25}2.38  & \cellcolor{gray!25}2.59   & \cellcolor{gray!25}0.90 & 6.46 
& \cellcolor{gray!25}3.03 & \cellcolor{gray!25}2.11  & \cellcolor{gray!25}2.25 & \cellcolor{gray!25}0.68 & \cellcolor{gray!25}7.08 \\

Corrected-Moment\tnote{1}(19 Color)  \cite{finlayson2013corrected}
&  2.96 & \cellcolor{gray!25}2.15  & \cellcolor{gray!25}2.37   & \cellcolor{gray!25}0.64 & 6.69 
& \cellcolor{gray!25}3.05 & 1.90  & 2.13 & \cellcolor{gray!25}0.65 & \cellcolor{gray!25}7.41 \\

Exemplar-based~\cite{joze2014exemplar}$^*$
  & 2.89 &  \cellcolor{gray!25}2.27  &\cellcolor{gray!25}2.42  & \cellcolor{gray!25}0.82 & 5.97 & --&-- &--& -- &-- \\
  
Chakrabarti \textit{et al.} 2015 ~\cite{chakrabarti2015color}
  & 2.56 &  1.67  & 1.89  &\cellcolor{gray!25} 0.52 & 6.07 & --&-- &--& -- &-- \\
  
Cheng \textit{et al.} 2015 \cite{cheng2015effective}
  & 2.42 &1.65 &1.75 & 0.38 & 5.87 &  2.18  & 1.48  & 1.64 & 0.46 & 5.03 \\
  
DS-Net (HypNet+SelNet) \cite{shi2016eccv}
  & 1.90  & 1.12 & 1.33 & 0.31 & 4.84 &  2.24  & 1.46 & 1.68 & 0.48 & 6.08 \\
  
CCC (dist+ext)  \cite{barron2015convolutional} 
  & 1.95 & 1.22 & 1.38 & 0.35 & 4.76 &  2.38 & 1.48 & 1.69 & 0.45 & 5.85 \\
FC$^{4}$  (AlexNet) \cite{hu2017cvpr}
& \textbf{1.77} & 1.11 & 1.29 & 0.34 & \textbf{4.29} &  2.12 & 1.53 & 1.67 & 0.48 & 4.78\\
FFCC  \cite{barron2017fourier}
  & 1.78 & \textbf{0.96} & \textbf{1.14} & \textbf{0.29} & 4.62 &  \textbf{1.99} & \textbf{1.31} & \textbf{1.43} & \textbf{0.35} & \textbf{4.75}\\
\hline

\textbf{Mean Shifted Gray Pixel}
& {3.45} &  \textbf{2.00} & \textbf{2.36} & \textbf{0.43} & 8.47 & \textbf{2.92} & \textbf{2.11} & \textbf{2.28} & \textbf{0.60} & 6.69 \\
\hline
\end{tabular}
}
}
\begin{flushleft}
\scriptsize
$^1$ For Correct-Moment \cite{finlayson2013corrected} we report reproduced and more detailed results by \cite{barron2015convolutional}, which slightly differs with the original results: {mean: 3.5, median: 2.6} for 19 colors and {mean: 2.8, median: 2.0} for 19 edges on Gehler-Shi Dataset.\\
$^*$ We mark Exemplar-based method with asterisk as it is trained and tested on a uncorrected-blacklevel dataset. 
\end{flushleft}
\end{threeparttable}
\begin{center}
\begin{threeparttable}
\subfloat[cross-dataset setting]{
\label{tab:agnostic}
\resizebox{\linewidth}{!}{

\begin{tabular}{l  rr rrr || rr rrr || rr }
\hline
Training set	&  \multicolumn{5}{c}{{NUS 8-Camera}}  & \multicolumn{5}{c}{{Gehler-Shi}}&\multicolumn{2}{c}{Average}\\
Testing set 	&  \multicolumn{5}{c}{{Gehler-Shi}}  & \multicolumn{5}{c}{{NUS 8-Camera}}&\multicolumn{2}{c}{runtime (s)}\\
    
				& { Mean} & { Median} & Trimean & Best 25\% & Worst 25\% &{ Mean} & { Median} & Trimean & Best 25\% & Worst 25\% & Train & Test\\
\hline
\multicolumn{5}{c}{{\em Learning-based Methods (agnostic-camera setting), Our rerun}} \\
Bayesian \cite{gehler2008bayesian} & 4.75 & 3.11 & 3.50 & 1.04 & 11.28 & 3.65 & 3.08 & 3.16 & 1.03 & 7.33 & 764 & 97\\
Chakrabarti \textit{et al.} 2015 ~\cite{chakrabarti2015color} Empirical&3.49 & 2.87 & 2.95 & 0.94 & {\bf 7.24} & 3.87 & 3.25 & 3.37 & 1.34 & 7.50 & -- & 0.30\\
Chakrabarti \textit{et al.} 2015 ~\cite{chakrabarti2015color} End2End&  3.52 & 2.71 & 2.80 & 0.86 & 7.72 & 3.89 & 3.10 & 3.26 & 1.17 & 7.95 & -- & 0.30\\

Cheng \textit{et al.} 2015 \cite{Chen-2015-CVPR}  &  5.52 &   4.52 & 4.79& 1.96 & 12.10   & 4.86  & 4.40  &  4.43& 1.72 & 8.87 & 245 & 0.25\\ 
FFCC \cite{barron2017fourier}  & 3.91 &  3.15 & 3.34 & 1.22 & {7.94} & 3.19 &  2.33 & 2.52 & 0.84 & 7.01 & 98 & 0.029\\ 
\hline
\multicolumn{5}{c}{{\em Physics-based Methods}} \\
IIC~\cite{tan2008color} & 13.62 &  13.56 & 13.45  & 9.46 & 17.98 & -- & -- & -- & -- & -- & -- &  --\\
Woo \textit{et al.} 2018~\cite{sung2018tip} & 4.30 &  2.86 & 3.31  & 0.71 & 10.14 & -- & -- & -- & -- & -- & -- &  --\\
\hline
\multicolumn{5}{c}{{\em Biological Methods}} \\
Double-Opponency~\cite{gao2015color} & 4.00 &  2.60 & --  & -- & -- & -- & -- & -- & -- & -- & -- &  --\\
ASM 2017~\cite{akbarinia2017colour} & 3.80 &  2.40 & 2.70  & -- & -- & -- & -- & -- & -- & -- & -- &  --\\
\hline
\multicolumn{5}{c}{{\em Learning-free Methods}} \\
White Patch \cite{brainard1986analysis}
& 7.55 &  5.68 & 6.35 & 1.45 & 16.12 & 9.91  & 7.44 &  8.78  & 1.44 & 21.27  & -- & 0.16\\
Grey World~\cite{buchsbaum1980spatial} & 6.36 &  6.28 & 6.28  & 2.33 & 10.58 & 4.59  & 3.46 &  3.81 & 1.16 & 9.85 & -- & 0.15\\
General GW~\cite{barnard2002comparison}
& 4.66  & 3.48 & 3.81 & 1.00 & 10.09 & 3.20 &  2.56 & 2.68 & 0.85 & 6.68& -- & 0.91\\ 
 2st-order grey-Edge ~\cite{van2007edge}
& 5.13 &4.44& 4.62 & 2.11 & 9.26  & 3.36  & 2.70  &2.80 & 0.89 & 7.14& -- & 1.30\\
1st-order grey-Edge ~\cite{van2007edge}

 & 5.33&4.52& 4.73 & 1.86 & 10.43 &  3.35  & 2.58 & 2.76 & 0.79 & 7.18& -- & 1.10\\
Shades-of-grey \cite{finlayson2004shades}
& 4.93 & 4.01 & 4.23 & 1.14 & 10.20 &  3.67  & 2.94 & 3.03 & 0.99 & 7.75& -- & 0.47\\
Grey Pixel (edge)~\cite{yang2015efficient} & 4.60 &  3.10 & --  & -- & -- & 3.15 & 2.20 & -- & -- & -- & -- &  0.88\\
LSRS  \cite{gao2014eccv}
&  3.31 & 2.80  & 2.87   & 1.14 & 6.39 
& 3.45 & 2.51  & 2.70 & 0.98 & 7.32 & -- & 2.60\\

Cheng \textit{et al.} 2014 \cite{Cheng14}
& 3.52 & 2.14& 2.47 & 0.50 & 8.74 & 2.93 & 2.33  & 2.42  & 0.78 & \textbf{ 6.13}& -- & 0.24\\

\textbf{Mean Shifted Gray Pixel}
& {3.45} &  {2.00} & {2.36} & {0.43} & 8.47
& {2.92} & {2.11} & {2.28} & {0.60} & 6.69 & -- & 1.32\\
\hline
\end{tabular} 
}
}
\end{threeparttable}
\end{center}
\end{table*}

\subsection{Camera-known Setting}
\label{sec:known} 
Camera-known setting (also termed as single-dataset setting) is the most common setting in related works, allowing extensive pre-training using a k-fold validation for learning-based methods. The results for this setting are summarized in table \ref{tab:maintable}b. Among all the compared methods, FFCC yields the best overall performance in both datasets. It is important to remark that, cross validation makes no difference in the performance of statistical methods. Therefore, in order to avoid repetition, the performance of competing statistical methods is not shown in this table (see next section). Remarkably, it is clear that, even in the known-camera setting, the proposed algorithm outperforms several learning-based methods (from Gamut \cite{gijsenij2010generalized} to the Exemplar-based method \cite{joze2014exemplar}) without extensive training and parameter tuning.

\subsection{Camera-agnostic Setting}
\label{sec:agnostic}
In order to allow for a fair comparison in the camera-agnostic scenario, learning-based methods should be re-trained for evaluation in the same conditions as statistical methods. Several state-of-the-art CNN-based methods are not publicly available. In this work, we were able to re-run the Bayesian method \cite{gehler2008bayesian}, Chakrabarti \textit{et al.}\cite{chakrabarti2015color}, FFCC \cite{barron2017cvpr}, and the method by Cheng \textit{et al.} 2015 \cite{cheng2015effective}, using the codes provided by the original authors. Note that
this list of methods includes FFCC, which showed the best overall performance in the camera-known setting. 

We train on one dataset and test on the other one. Both datasets share no common cameras, thus meeting our requirement of being ``camera-agnostic''. For the results reported in this section, we use the best or final setting for each method: Bayes (GT) for Bayesian; Empirical and End-to-End training for Chakrabarti \textit{et al.} ~\cite{chakrabarti2015color}; $30$ regression trees for Cheng \textit{et al.}; full image resolution and $2$ channels for FFCC\footnote{Scripts for re-running these methods will also be public.}. Obtained results are summarized in Table~\ref{tab:maintable}a.

Obtained results are summarized in Table~\ref{tab:maintable}a. From this table, it is clear that the proposed MSGP algorithm outperforms both learning-based and statistical methods. Except FFCC, selected learning-based methods perform relatively worse in camera-agnostic setting, as compared to statistical methods. Due to their nature, it is not surprising that learning-based methods degrade in their performance in the camera-agnostic scenario. However, the fact that learning-based methods are outperformed by statistical methods is an interesting finding. On one side, if we use learning-based methods trained for a given dataset or "a bag of camera models", we may fail in the camera-agnostic setting. In contrast, in the both camera-agnostic/known setting, the proposed statistical method provides stable performance.

\subsection{Algorithm parameters}
\label{sec:params}

\noindent \textbf{The role of bandwidth $h$.} 
The bandwidth $h$ determines the domain size where Mean Shift computes the pixel divergence. Here we evaluate variants of the proposed method by changing $h$ to be $1\mathrm{e}{-4}$, $1\mathrm{e}{-3}$ and $1\mathrm{e}{-2}$. Table \ref{tab:testK} shows that the bandwidth $1\mathrm{e}{-3}$ gives a good trade-off between mean and median error on two datasets. For reference purposes, Table \ref{tab:testK} also includes performance results obtained when the distance function in Eq. \ref{eq:kernel} uses only angular information in $D(\cdot)$.

\begin{table}
\small
\caption{Comparison between Mean Shift Clustering vs. K-means Clustering in our task. ``angle'' refers to using angular distance only in Mean Shift instead of the proposed hybrid distance.} 
\label{tab:testK}
\begin{center}
  \resizebox{1.0\linewidth}{!}{
  \begin{tabular}{l|r|r|r|r|r|r}
  \hline
  &  \multicolumn{3}{c}{{SFU Color Checker}}  & \multicolumn{3}{c}{{NUS 8-Camera}}\\
  & Mean & Med & Trimean & Mean & Med & Trimean\\
  \hline
 \multicolumn{1}{c}{{\em Mean Shift}} \\
h=$1\mathrm{e}{-3}$ (angle)  & 3.62 &  2.08 &  2.42  & 3.00 & 2.10 & 2.26  \\ 
h=$1\mathrm{e}{-4}$ & 3.51 &  2.04 & 2.38  & 3.32 & 2.13 & 2.39 \\
h=$1\mathrm{e}{-3}$  & 3.45 &  2.00 &  2.36  & 2.92 & 2.11 & 2.28  \\ 
h=$1\mathrm{e}{-2}$ & 3.48 &  2.11 & 2.44 & 3.00 & 2.19 & 2.39 \\ 
  \hline
 \multicolumn{1}{c}{{\em Kmeans}} \\
K=2 & 3.75 &  2.18 & 2.54  & 3.00 & 2.10 & 2.28 \\
K=5  & 4.44 &  2.46 & 2.73  & 3.32 & 2.13 & 2.37  \\ 
K=9  & 4.50 &  2.51 & 2.80  & 3.37 & 2.19 & 2.39 \\ 
  \hline
  \end{tabular}
  }
\end{center}
\end{table}

\noindent \textbf{Clustering Algorithm} We compare two clustering methods, Mean Shift and K-means\footnote{We use clustering to find the mode \textit{i.e.} the dominating illumination color, while we don't need all all clustered indexes. We note that other clustering methods (\textit{e.g.} spectral clustering) may work well. We selected Mean Shift due to its fast computation and robustness to the outliers. }. Here we evaluate variants of K-means by changing the number of clusters $K$ to $2$, $5$ and $9$. Table~\ref{tab:testK} shows that, in general, MS gives better results. This can be attributed to the fact that Mean Shift is more robust to outliers than K-means. Among all K-means invariants, the $2$-cluster setting performs best. This suggests that $S$ usually contains $1-2$ elongated clusters. 

\begin{figure}[ht]
\begin{center}
\subfloat[$15.18^\circ$]{\includegraphics[width=.28\linewidth]{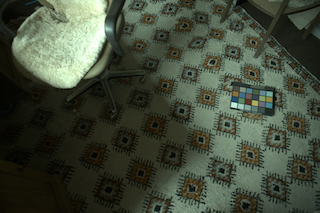}\label{fig:no_cc_43}}\hspace{.5cm}
\subfloat[$26.13^\circ$]{\includegraphics[width=.28\linewidth]{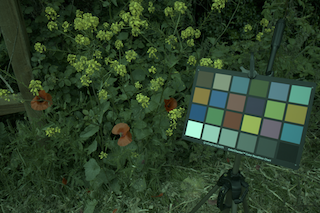}\label{fig:no_cc_53}}\\

\subfloat[$10.00^\circ$]{\includegraphics[width=.28\linewidth]{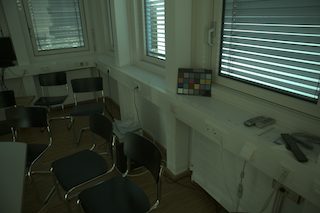}\label{fig:2illu_cc_9}}\hspace{0.5cm}
\subfloat[$15.77^\circ$]{\includegraphics[width=.28\linewidth]{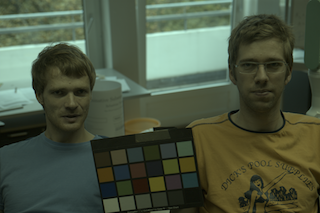}\label{fig:2illu_cc_25}}

\caption{Example failure cases with their angular errors. (a,b) are examples with no detectable gray pixels (note that the ground truth color chart is masked in evaluation). (c,d) are examples with mixed illumination: indoor illumination and outdoor illumination.}
\label{fig:failcases}
\end{center}
\end{figure} 

\section{\uppercase{Limitations and Conclusions}}

\noindent Our method relies on gray pixels and their statistics for one global illumination estimation. Therefore, in some extreme cases, when there are no detectable gray pixels or there are gray pixels representing two not-same-color illuminations, our method fails. In Figure \ref{fig:failcases}, two no-gray-pixel examples and two double-illumination examples are shown. Cheng \textit{et al.} \cite{cheng2016cvpr} claimed that in SFU Color Checker Dataset \cite{shi2010re}, there are $66$ two-illumination images (image list released). It is worthy to mention that the images where we fail overlap largely with this two-illumination list. As mixed-illumination problem is a different task and out of the scope of this paper, we refer readers to \cite{cheng2016cvpr} for details.

In this paper, we presented a statistical method for tackling the problem of color constancy. The proposed method relies on gray pixel detection and mean shift clustering in order to estimate the illumination of the scene based on the statistical properties of the gray pixels of the input image. In the camera-agnostic scenario, in which color constancy is to be applied to images captured with unknown cameras, the proposed method outperforms both learning-based and statistical state-of-the-arts.

The proposed method is easy to implement, training-free, and depends only on two parameters, namely the percentage of gray pixels $N\%$ and the Mean Shift bandwidth $h$. With our method, processing a $2000\times1500$ linear RGB image takes about 1.32 seconds with unoptimized MATLAB code running in a CPU Intel i7 2.5 GHz. The method can be adapted to other color spaces (\textit{e.g.} Lab) without any performance drop.

\bibliographystyle{apalike}
{\small \bibliography{vissap2019}}
\section*{\uppercase{Appendix}}
\subsection*{Detailed settings of learning-based methods}
\label{sec:detailed_setting}

To evaluate the performance of learning-based method in camera-agnostic scenario, we re-run the Bayesian method \cite{gehler2008bayesian}, Chakrabarti \textit{et al.} 2015~\cite{chakrabarti2015color}, FFCC \cite{barron2017cvpr}, and the method by Cheng \textit{et al.} 2015 \cite{cheng2015effective}, using the codes provided by the authors. FFCC shows the best overall performance in the camera-known setting. Our experimental settings for re-running the aforementioned algorithms are summarized below:

\vspace{0.5cm}

\begin{tabular}{p{0.4\linewidth}p{0.5\linewidth}}
\textbf{Bayesian method \cite{gehler2008bayesian}} & Among all variations of Bayesian methods stated in \cite{gehler2008bayesian}, we use Bayes (GT) but without indoor/outdoor split, to which Bayes (tanh) is sensible. The ground truth of training illuminations (\textit{e.g.} Gehler-Shi) is used as point-set prior for testing on the other dataset (\textit{e.g.} NUS 8-camera)\\
\textbf{Chakrabarti \textit{et al.} 2015~\cite{chakrabarti2015color}} & We use both variations given by the author: the empirical and the end-to-end trained method. We keeps all training hyperparameters same, \textit{e.g.} epoch number, momentum and learning-rate for SGD.\\
\textbf{FFCC \cite{barron2017cvpr}} & For fair comparison, we use Model (J) (FFCC full,4 channels) in \cite{barron2017cvpr}, which is free of camera metadata and semantic information but still state-of-the-art.\\
\textbf{Cheng \textit{et al.} 2015 \cite{cheng2015effective}} & Same as \cite{cheng2015effective}, we use four 2D features with an ensemble of regression trees (K=30).
\end{tabular}

\vfill
\end{document}